\documentclass{article}

\usepackage{microtype}
\usepackage{graphicx}
\usepackage{subfigure}
\usepackage{booktabs} % for professional tables

\usepackage{hyperref}

\usepackage[accepted]{icml2024}

\usepackage{amsmath}
\usepackage{amssymb}
\usepackage{mathtools}
\usepackage{amsthm}

\usepackage[capitalize,noabbrev]{cleveref}

\theoremstyle{plain}
\newtheorem{theorem}{Theorem}[section]

\theoremstyle{definition}
\newtheorem{definition}[theorem]{Definition}

\theoremstyle{remark}

\usepackage{bm}

\usepackage{xcolor}  
\newcommand{\fix}[1]{{#1}}

\usepackage[textsize=tiny]{todonotes}

\icmltitlerunning{Offline Multi-Objective Optimization}

\begin{document}

\twocolumn[
\icmltitle{Offline Multi-Objective Optimization}



\icmlsetsymbol{equal}{*}

\begin{icmlauthorlist}
\icmlauthor{Ke Xue}{equal,nju,nju2}
\icmlauthor{Rong-Xi Tan}{equal,nju,nju2}
\icmlauthor{Xiaobin Huang}{nju,nju2}
\icmlauthor{Chao Qian}{nju,nju2}
\end{icmlauthorlist}

\icmlaffiliation{nju}{National Key Laboratory for Novel Software Technology, Nanjing University, China}
\icmlaffiliation{nju2}{School of Artificial Intelligence, Nanjing University, China}

\icmlcorrespondingauthor{Chao Qian}{qianc@nju.edu.cn}

\icmlkeywords{black-box optimization, offline optimization, multi-objective optimization, Bayesian optimization}

\vskip 0.3in
]



\printAffiliationsAndNotice{\icmlEqualContribution} % otherwise use the standard text.

\begin{abstract}
Offline optimization aims to maximize a black-box objective function with a static dataset and has wide applications. In addition to the objective function being black-box and expensive to evaluate, numerous complex real-world problems entail optimizing multiple conflicting objectives, i.e., multi-objective optimization (MOO). Nevertheless, offline MOO has not progressed as much as offline single-objective optimization (SOO), mainly due to the lack of benchmarks like Design-Bench for SOO. To bridge this gap, we propose a first benchmark for offline MOO, covering a range of problems from synthetic to real-world tasks. This benchmark provides tasks, datasets, and open-source examples, which can serve as a foundation for method comparisons and advancements in offline MOO. Furthermore, we analyze how the current related methods can be adapted to offline MOO from four fundamental perspectives, including data, model architecture, learning algorithm, and search algorithm. Empirical results show improvements over the best value of the training set, demonstrating the effectiveness of offline MOO methods. As no particular method stands out significantly, there is still an open challenge in further enhancing the effectiveness of offline MOO. We finally discuss future challenges for offline MOO, with the hope of shedding some light on this emerging field. Our code is available at \url{https://github.com/lamda-bbo/offline-moo}.
\end{abstract}

\section{Introduction}
Creating new designs to optimize specific properties is a widespread challenge, encompassing various domains such as real-world engineering design~\cite{RE}, protein design~\cite{antBO}, and molecule design~\cite{lambo}. 
Many methods generate new designs by iteratively querying an unknown objective function that maps a design to its property score. 
However, in real-world situations, evaluating the objective function can be time-consuming, costly, or even dangerous~\cite{drug}.
To optimize for the next candidate design based on accumulated data, a rational approach prefers to build a model, use it to guide the search, and select a suitable candidate for the evaluation. This approach is known as offline model-based optimization~\cite{design-bench}.

Offline model-based optimization solely permits access to an offline dataset and does not permit iterative online evaluation (i.e., only one batch of real evaluations), which presents notable challenges in comparison to more commonly studied online optimization.
A common approach is to train a deep neural network (DNN) model $f_{\theta}(\cdot)$ on a static dataset and use the trained DNN as a proxy (also known as a surrogate model). The DNN proxy enables gradient descent on existing designs, which can result in an improved solution that is even better than the previously seen best one sometimes. 
However, this approach has a drawback: the trained proxy is prone to out-of-distribution problems, i.e., it makes inaccurate predictions when applied to data points that deviate significantly from the training distribution. 
Besides, in some cases, the learned proxy has a non-smooth landscape, posing challenges to optimize in it.
\fix{Many recent studies try to address these issues from different perspectives, e.g., COMs~\cite{coms} uses adversarial training to create a smooth proxy;
RoMA~\cite{ROMA} employs a local smoothness prior to alleviate the fragility of the proxy and achieves robust estimation by model adaptation;
Tri-Mentoring~\cite{tri-mentoring} effectively utilizes weak ranking supervision signals among proxies and achieves a robust ensemble of proxies by an adaptive soft-labeling module; just to name a few.}

In addition to the objective function being black-box and the evaluations being costly, numerous complex real-world problems entail optimizing multiple objectives, frequently with conflicting requirements, which can be formulated as multi-objective optimization (MOO) problems~\cite{mo-book,mo-book2}. The goal of MOO is to find a set of solutions that represent the optimal trade-offs among the various objectives, thereby significantly augmenting the complexity of the problem compared to single-objective optimization (SOO) which aims to obtain a single optimal solution. 
Indeed, MOO is a more prevalent problem than SOO. Many single-objective problems are essentially multi-objective in nature, but they are often converted into a single objective by assigning weights to multiple objectives, primarily due to the challenges associated with solving MOO~\cite{lambo,harmful}.

Recently, researchers have recognized the significance of directly modeling MOO problems~\cite{nsgaii,qparego}. The demand for offline MOO is also gradually increasing. However, the progress of offline MOO is far behind compared to offline SOO. Thanks to the remarkable benchmark Design-Bench~\cite{design-bench}, \fix{several advanced offline SOO algorithms have been proposed, which can perform well even in high-dimensional and complex search spaces~\cite{bdi,iom,ict,tri-mentoring,ddom,bootgen,pgs,leo,uehara2024bridging}}. Unfortunately, there has been no such benchmark available for offline MOO, which hinders its progress. Even for online MOO, most works conduct evaluations on synthetic functions with a few exceptions that include real-world applications. This calls for a much-needed push towards more challenging benchmarks for reliable evaluation of MOO, especially in the offline setting. 

\begin{figure}[t!]\centering
\includegraphics[width=0.95\linewidth]{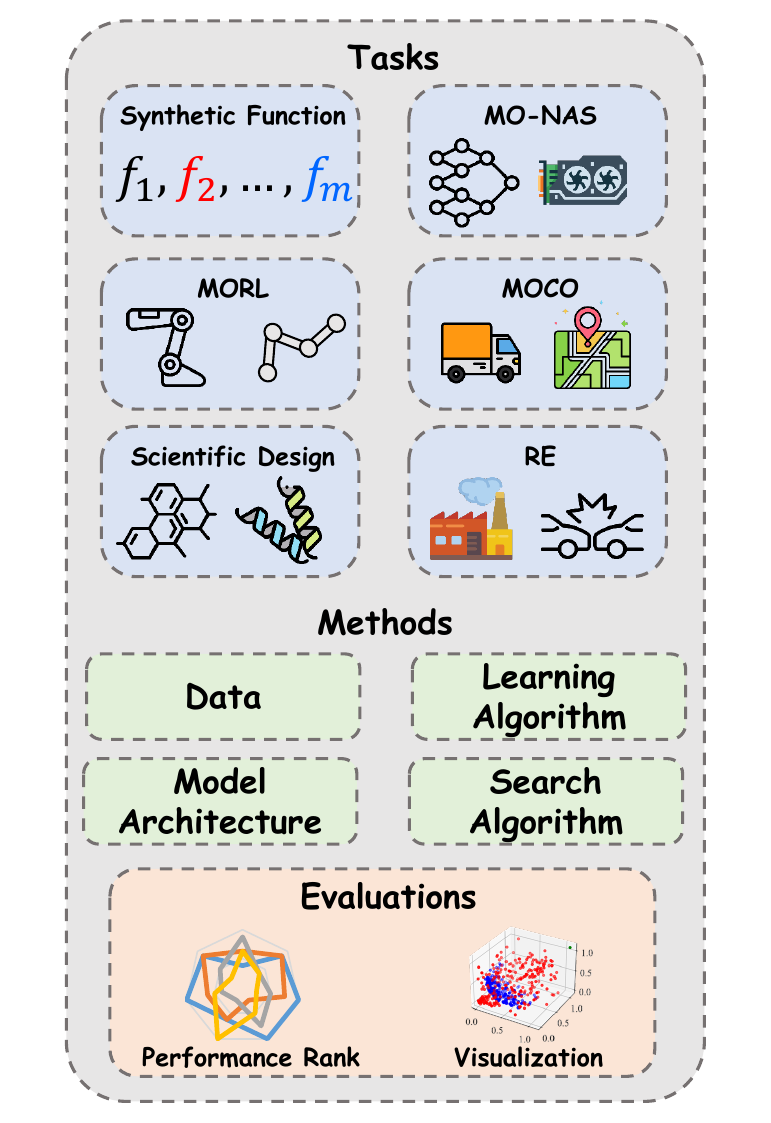}
\vspace{-0.5em}
\caption{Benchmarks for offline MOO.}\label{fig:overview}
\vspace{-1.5em}
\end{figure}

In this paper, we propose a first benchmark for offline MOO, where the tasks range from synthetic functions to real-world science and engineering problems, as shown in Figure~\ref{fig:overview}. To facilitate future research, we release our benchmark tasks and datasets with a comprehensive evaluation of different approaches and open-source examples. Specifically, we analyze an offline MOO method from four fundamental perspectives including data, model architecture, learning algorithm, and search algorithm, and propose two types of potential methods, i.e., DNN-based and Gaussian process-based offline MOO, by learning techniques from related areas such as offline SOO and multi-objective Bayesian optimization. Experimental results show that the proposed methods can achieve better results than the optimal ones in the training set. However, as no single method stands out significantly, how to enhance the effectiveness of offline MOO remains open. Our work serves as a starting point for offline MOO, and we hope it can encourage more explorations in this emerging area.

Our contributions can be summarized as follows:\vspace{-0.6em}
\begin{itemize}
    \item We propose a first benchmark for offline MOO, providing not only a large amount of offline data but also commonly used MOO interfaces. This facilitates the integration of a wider range of problems and algorithms.\vspace{-0.3em}
    \item We analyze an offline MOO method from four fundamental perspectives, including data, model architecture, learning algorithm, and search algorithm, and compare various implementations within a unified framework, making it convenient for researchers to compare their performance in a clear manner.\vspace{-0.3em}
    \item We provide extensive empirical studies, and also discuss challenges and future directions of offline MOO.
\end{itemize}

\section{Background}
\subsection{Offline Optimization}
Given an offline-collected static dataset $\mathcal{D} = \left \{(\bm x_i, y_i) \right \}_{i=1}^N$, offline model-based optimization aims to find an optimal solution (also called ``design" in many scenarios) $\bm x^*$ that minimizes the black-box objective function $f(\cdot)$, i.e., $\bm x^* = \arg \min_{\bm x \in \mathcal X} f(\bm x)$.

A common approach for solving offline optimization problems is approximating the black-box objective function $f(\cdot)$ using a surrogate model, e.g., DNN. The parameters of DNN can be trained by minimizing the mean squared error between the predictions and the true scores. After that, the trained DNN model is used as a surrogate evaluator to optimize using a search algorithm, e.g., gradient descent. 

Since offline optimization does not allow iterative real evaluations, the algorithm is expected to output a proper solution that is better than the best solution seen in the dataset. However, in practice, producing a single better design entirely from offline data is very difficult, so offline optimization methods are more commonly evaluated in terms of ``$P$ percentile of top $K$" performance~\cite{MINs}, where the algorithm produces $K$ candidates and the $P$ percentile objective value determines the final performance.

Many real-world tasks are inherently multi-objective, but they are usually simplified and formulated as single-objective problems. 
For example, neural architecture search (NAS) should not only maximize accuracy, but also minimize the scale of the model~\cite{mo-nas}; protein design should take efficacy, toxicity, and yield into consideration simultaneously~\cite{lambo}. 
In this paper, we aim to highlight the importance and challenges of offline MOO, and provide a benchmark and comprehensive empirical studies on it.  

\subsection{Multi-Objective Optimization}\label{sec:moea}
First, we give a brief introduction to multi-objective optimization problems, which can be defined as
\begin{equation}\label{eq:moo}
    \min\nolimits_{\bm x\in \mathcal{X}} \bm{f}(\bm{x}) = (f_1(\bm{x}), \dots, f_m(\bm{x})),
\end{equation}
where $\bm{x} = (x_1,\dots,x_D)$ is a solution, $\bm f: \mathcal{X} \to \mathbb R^m$ constitutes $m$ objective functions, $\mathcal{X}$ is the solution space, and $\mathbb R^m$ is the objective space. For a non-trivial problem, no single solution can optimize all objectives at the same time, and we have to make a trade-off among them~\cite{recombination,sto-population}. 
\begin{definition}
A solution $\bm x^*$ is Pareto-optimal with respect to Eq.~(\ref{eq:moo}), if $\nexists \bm x\in \mathcal{X}$ such that $\forall i: f_i(\bm{x})\leq f_i(\bm{x}^*)$ and $\exists i: f_i(\bm{x})< f_i(\bm{x}^*)$. The set of all Pareto-optimal solutions is called Pareto-optimal set (PS). The set of the corresponding objective vectors of PS, i.e., $\{\bm{f}(\bm{x})\mid\bm{x}\in \text{PS}\}$, is called Pareto front (PF).
\end{definition}
Instead of focusing on a single optimal solution in SOO, the goal of MOO is to find \emph{a set of solutions} that can approximate the PF well. Next, we will briefly introduce two main kinds of methods for solving MOO problems. 

\textbf{Multi-objective evolutionary algorithm (MOEA).}
Evolutionary algorithms~\cite{eabook,elbook} have demonstrated their effectiveness in solving MOO problems. MOEA follows the population-based search by iterative parent selection, reproduction, and survivor selection, which can approximate the Pareto optimal solutions within one execution, with each solution in the population representing a unique trade-off among the objectives~\cite{moea-book}. Over the last decades, there have been a lot of well-known MOEAs developed~\cite{coello2007evolutionary}. NSGA-II~\cite{nsgaii} is a typical Pareto dominance-based MOEA, using fast non-dominated sorting for selecting solutions. 
MOEA/D~\citep{moead} is a decomposition-based MOEA, converting an MOO problem into multiple SOO sub-problems through a number of weights, where neighboring solutions work cooperatively for the optimal solutions of the single-objective sub-problems. NSGA-III~\cite{deb2013evolutionary} is proposed to handle MOO problems with many objectives (having four or more objectives), by using reference points to assist the selection within non-dominated solutions.

\textbf{Multi-objective Bayesian optimization (MOBO).} 
Many real-world MOO tasks are expensive to evaluate. MOBO is suitable for these tasks due to its high sample-efficiency. Based on the observed data, MOBO learns a surrogate model, e.g., Gaussian process (GP)~\cite{gpml}, searches for new promising candidates based on an acquisition function built on the surrogate model, and queries the quality of these candidates with the ground truth black-box objectives. Existing MOBO methods mainly fall into the following three types. Hypervolume based methods consider the widely-used hypervolume metrics in acquisition function~\cite{ehvi1,dgemo,qnehvi,hvkg}. Scalaraization based methods reduce the MO acquisition function into one or multiple SO problems via scalarization~\cite{parego,moeadego,mobors,hvrs}. Information-theoretic methods select points to reduce the uncertainty of the unknown Pareto front~\cite{pesmo, pfes, jes, pf2es}. Besides these methods, there are also works addressing MOBO in other scenarios, such as high-dimensional space~\cite{lamoo} and sequence space~\cite{lambo}.

While MOO has made significant progress, most existing methods either use handcrafted mechanism and lack a learning mechanism (e.g., MOEA) or are unable to leverage a large amount of offline data for scalable learning (e.g., MOBO), restricting their applications in offline MOO tasks. Additionally, there is a lack of benchmark for offline MOO. 
Note that a good benchmark plays a crucial role in the advancement of a research field and the development of state-of-the-art algorithms, such as NASBench~\cite{nas101} and HPO-B~\cite{hpob} for BBO, D4RL~\cite{d4rl} and NeoRL~\cite{neorl} for offline RL, and Design-bench~\cite{design-bench} for offline SOO. In the following, we will propose the problem of offline MOO and provide a large-scale benchmark, covering a wide range of tasks and methods.

\section{Offline MOO Benchmark}
\fix{We present the problem formulation in Section~\ref{sec:3.1} and the process of collecting the dataset for our Offline MOO Benchmark (Off-MOO-Bench) in Section~\ref{sec:data-collect}. 
We will introduce the tasks and methods in our benchmark in Sections~\ref{sec:tasks} and \ref{sec:method}, respectively.}

\subsection{Offline MOO}\label{sec:3.1}
Given an offline-collected static dataset $\mathcal{D} = \left \{(\bm x_i, \bm y_i) \right \}_{i=1}^N$, where $\bm x_i$ and $\bm y_i$ denotes a solution and its objective vector, respectively, offline MOO aims to find a set of solutions to approximate the Pareto front of the MOO problem in Eq.~(\ref{eq:moo}). Similar to offline SOO, offline MOO only allows access to the offline dataset and does not permit iterative online evaluation. Besides, the MOO nature makes offline MOO more challenging. 

\fix{Due to the goal of finding a set of solutions rather than a single solution, the commonly used measure ``$P$ percentile of top $K$" in offline SOO cannot be directly applied for offline MOO. 
In our experiments, each offline MOO algorithm first outputs a certain number of solutions (e.g., 256 and 32) to be evaluated. To report the ``$P$ percentile" measure, we use the NSGA-II selection procedure (i.e., first applying non-dominated sorting then selecting the top solutions)~\cite{nsgaii} to eliminate the top $1-P\%$ of solutions and report the remaining solutions' metrics as the evaluation results.} 
There are two commonly used metrics in MOO, i.e., inverted generational distance (IGD)~\citep{igd}, which measures the distance between a solution set and the true Pareto front, and hypervolume (HV)~\cite{hv}, which measures the volume of the objective space between a reference point and the objective vectors of a solution set, reflecting both convergence and diversity of the solution set. Because the calculation of IGD requires knowing the true Pareto front, which cannot be obtained in real-world tasks, we use HV as the metric in our benchmark. The reference point required to calculate HV is set to the nadir point, each dimension of which corresponds to the worst value of one objective. Details are provided in Appendix~\ref{app:settings}.

\subsection{Dataset Collection}\label{sec:data-collect}

We use three representative MOEAs, i.e., NSGA-II, MOEA/D, and NSGA-III, introduced in Section~\ref{sec:moea} to collect the data for all the tasks. For each problem, we run these three expert algorithms independently and collect the data as our dataset.
However, only using the expert algorithms may result in a significant difference between our data distribution and diverse reality distribution. Thus, we introduce a probability of accepting inferior solutions during the survivor selection process of the expert algorithms.
In addition, to solve problems with different search spaces, we also employ various types of evolutionary operators. Detailed settings are provided in Appendix~\ref{app:settings}.

The complex objective space of real-world problems presents significant challenges for offline MOO. We provide the visualizations of the objective space in Appendix~\ref{app:results}. Compared to Design-Bench for offline SOO, our benchmark includes more data due to the inherent challenge of MOO. Additionally, our framework presents many easy-to-use interfaces to facilitate the integration with other algorithm implementations, including sub-problem generation, weight decomposition, HV evaluation, etc.

\section{Tasks}\label{sec:tasks}
In this section, we describe the set of tasks included in our benchmark. An overview of the tasks is provided in Table~\ref{tab:tasks}. Each task in our benchmark suite comes with a dataset $\mathcal{D}$, along with a ground-truth oracle objective function $\bm{f}$ that can be used for evaluation. An offline algorithm should not query the ground-truth oracle function during training, even for hyperparameter tuning. We first discuss the tasks in our benchmark.
Detailed information about these tasks are provided in Appendix~\ref{app:task} due to space limitation.

\begin{table*}[htbp]
\caption{Properties of the tasks in offline MOO Benchmark.}\vspace{0.3em}
\centering
\begin{tabular}{c|cccc}
\toprule
Task Name              & Dataset size & Dimensions & \# Objectives & Search space        \\ \midrule
Synthetic Function     &  60000    &  2-30          & 2-3          & Continuous  \\
MO-NAS                 &  9735-60000    &   5-34         & 2-3            & Categorical  \\
MO-Swimmer             &   8571   &  9734          & 2            & Continuous  \\
MO-Hopper              &   4500   &   10184         & 2            & Continuous  \\
MO-TSP                 &   60000   &  20-500          & 2-3            & Permutation \\
MO-CVRP                &  60000    &  20-100          & 2-3            & Permutation \\
MO-KP                  &   60000   & 50-200           & 2-3           & Permutation \\
MO-Portfolio                  &   60000   & 20           & 2          & Continuous \\
Molecule   &  49001    &      32     & 3            & Continuous    \\
Regex   &  42048   &      4     & 2            & Sequence    \\
RFP     & 4937 &     4     & 2            & Sequence  \\ 
ZINC     & 48000 &     4     & 2            & Sequence  \\ 
Real-world Application & 60000     &  3-6          & 2-6          & Continuous \& Mixed  \\
\bottomrule 
\end{tabular}\label{tab:tasks}
\vspace{-1em}
\end{table*}

\subsection{Synthetic Function}
We first use various synthetic functions as our tasks, which encompass several popular MOO problem sets, i.e., DTLZ~\cite{dtlz}, ZDT~\cite{zdt}, Omni-test~\cite{omnitest}, and VLMOP~\cite{vlmop}. The search space is continuous, and the objectives are predetermined by the function designers.
Although these synthetic functions may not be considered ``realistic", they possess certain advantages and are worth considering for the following reasons:
a) Their analytical expressions are known, allowing us to obtain the actual Pareto front for better understanding the problem's characteristic and the algorithm's behavior;
b) They can be easily configured to any input dimension and any number of objectives, making them suitable for testing large-scale and many-objective optimization algorithms;
c) They are computationally efficient to evaluate, enabling us to collect lots of data and assess the scalability of offline MOO algorithms.
We implement these synthetic functions and collect the data.

\subsection{Multi-Objective Neural Architecture Search}
NAS has paved a promising path towards alleviating the unsustainable process of designing DNN architectures by automating the pipeline. Apart from the prediction error, recent NAS works also consider other objectives, e.g., the number of parameters. These NAS tasks are intrinsically MOO problems, aiming to achieve trade-offs of the multiple design criteria~\cite{mo-nas}.
\fix{We provide a toy example named NAS-Bench-201-Test, which uses a categorical cell-based search space~\cite{nas201}. Besides, C-10/MOP and IN-1K/MOP from~\citet{mo-nas} are also included, where both \textit{micro} and \textit{macro} search spaces are used. 
For these tasks, there are three objectives to be minimized, i.e., error, number of parameters and edge GPU latency, which measure the model's performance, scale, and  the GPU's efficiency during model execution, respectively. The data is from~\citet{mo-nas}. Detailed information about tasks is provided in Appendix~\ref{sec:mo-nas}.}

\subsection{Multi-Objective Reinforcement Learning}
Decision making in practical applications usually involves reasoning about multiple, often conflicting, objectives~\cite{d4morl}. For example, when designing a control policy for a running quadruped robot, we need to consider two conflicting objectives: running speed and energy efficiency. Multi-objective reinforcement learning (MORL) aims to learn agents that can handle such a challenging task. 
We consider two locomotion tasks in the popular MORL benchmark MuJoCo~\cite{mujoco}, i.e., MO-Swimmer and MO-Hopper. Their search space is the parameters of an agent, which is much larger than other tasks.
The two objectives in MO-Swimmer are speed and energy efficiency, and MO-Hopper considers two objectives related to running and jumping. The data is collected by us \fix{via running PG-MORL~\cite{pgmorl}}.

\subsection{Multi-Objective Combinatorial Optimization}
Multi-objective combinatorial optimization (MOCO) commonly exists in industries, such as transportation, manufacturing, energy, and telecommunication~\cite{nhde}.
We consider three typical MOCO problems that are commonly studied, i.e., multi-objective traveling salesman problem (MO-TSP), multi-objective capacitated vehicle routing problem (MO-CVRP), and multi-objective knapsack problem (MO-KP)\fix{, and a multi-objective portfolio allocation (MO-Portfolio) problem}. \textbf{MO-TSP} has $n$ nodes, where each node has two sets of 2-dimensional coordinates. There are two objectives, each of which corresponds to the travel cost calculated using one set of 2-dimensional coordinates of all nodes. \textbf{MO-CVRP} has $n$ customer nodes and a depot node, with each node featured by a 2-dimensional coordinate and each customer node associated with a demand. Following the common practice, we consider two objectives, i.e., the total tour length and the longest length of the route. \textbf{MO-KP} has $n$ items, with each taking a weight and two separate values. The goal is to maximize the sum of their 2-dimensional objective vectors (corresponding to two objectives) under the constraint that the sum of weights does not exceed a capacity. The search space of these problems is a permutation space, and we use the corresponding operator in MOEA to search in it. \fix{The \textbf{MO-Portfolio} task is continuous and it is based on the Markowitz Mean-Variance Portfolio Theory~\cite{portfolio}, where the two objectives, i.e., expected returns and variance of returns, are used to illustrate the relations between beliefs and choice of portfolio.} The data is collected by us.

\subsection{Scientific Design}
Many real-world scientific problems also involve MOO. We consider molecule design and protein design, which are two important sequence optimization problems. The data of these tasks is collected by us.

\textbf{Molecule design} is critical to pharmaceutical drug discovery~\cite{drug}. 
Previous research has typically required the generated molecules to fulfill several objectives, e.g., new drugs should generally be non-toxic and ideally easy-to-synthesize, in addition to their primary purpose. In this task, we consider two objectives based on prior work in molecular design~\cite{lamoo}, i.e., activity against biological targets GSK3$\beta$ and JNK3, respectively. The solution is optimized in a pretrained 32-dimensional continuous latent space~\cite{motifs}, which is then decoded into molecular strings and fed into the property evaluators. 

\textbf{Protein design} is the process of creating new or improved protein structures for use as biomarkers, therapeutics, etc. We consider the following three tasks. \textbf{Regex} is a basic task (around 32 tokens) for protein design, where the objectives are to maximize the counts of multiple bigrams. \fix{\textbf{ZINC} is a small scale task (around 128 tokens) to optimize the chemical properties of a small molecule. The two objectives are to maximizing the logP (the octanol-water partition coefficient) and QED (quantitative estimate of druglikeness).} \textbf{RFP} is a large-scale task (around 200 tokens) designed to simulate searching for improved red fluorescent protein (RFP) variants, a problem of significant interest to biomedical researchers. The two objectives are maximizing the solvent-accessible surface area and the stability of RFP, respectively.

\subsection{Real-World Application}
MOO has applications in many real-world tasks. We select several real-world multi-objective engineering design problems from RE suite~\cite{RE}, including four bar truss design, pressure vessel design, disc brake design, vehicle crashworthiness design, rocket injector design, etc. These tasks provide various challenges for offline MOO, e.g., they have different number of objectives and different types of variables. We use the evaluation interface from RE and collect the data ourselves.

\section{Offline MOO Method}~\label{sec:method}
\hspace{-1em} This section introduces the approaches for offline MOO. Though no specific approach has yet been developed to address offline MOO problems, we can adapt the techniques from other related topics, such as offline SOO, MOBO, and surrogate-assisted evolutionary algorithm (SAEA)~\cite{saea}. All of these methods use a surrogate model and conduct searches within it. Offline SOO uses a neural network to build a surrogate model, while MOBO typically uses a Gaussian process (GP). SAEA may use both, with a focus on \emph{``How to properly use the surrogate during the iterative search process"}, which, however, is not consistent with offline settings that do not support iterative search. As a result, we consider modifying offline SOO and MOBO methods to address offline MOO tasks, by using the DNN-based and GP-based surrogate models, respectively. We will consider four fundamental components of an offline MOO method: data, model architecture, learning algorithm, and search algorithm, which are shown below.

\subsection{DNN-Based Offline MOO Method}~\label{sec:5.1}
\hspace{-0.8em} DNN-based methods~\cite{ROMA,coms,design-bench,tri-mentoring} have shown impressive performance in offline SOO due to its ability to learn from a large amount of historical data, while also being able to perform search using gradient ascent within it. Model architecture design is a key aspect in this kind of method, especially for offline MOO. We consider the following three models.

\textbf{End-to-end model} is a straightforward approach, using a DNN to learn an approximation of $m$ objectives simultaneously, where the model takes $\bm{x}$ as input and outputs an $m$-dimensional objective vector directly.

\textbf{Multiple models} maintains $m$ independent surrogate models for an $m$-objective problem, which is a common practice in MOBO. Each individual model learns an objective function independently, allowing for the natural use of offline SOO techniques such as COMs~\cite{coms}\fix{, RoMA~\cite{ROMA}, IOM~\cite{iom}, ICT~\cite{ict}, and Tri-Mentoring~\cite{tri-mentoring}}.

\textbf{Multi-head models.}
We observe that learning multiple objective functions simultaneously is similar to multi-task learning (MTL)~\cite{mtl-survey}, whose aim is to leverage useful information contained in multiple related tasks to help improve the generalization performance of all the tasks. As a commonly used model in MTL, multi-head models can also serve as a fundamental model for offline MOO. Furthermore, we propose utilizing training techniques (e.g., GradNorm~\cite{gradnorm} and PcGrad~\cite{gradient-surgery}) from MTL to assist model training of offline MOO.

\textbf{Data pruning.} During the training process, we find that using all the data for model training results in a significant inferior performance: the search algorithm can only obtain few solutions. This phenomenon occurs across all model structures, which may be attributed to the training data quality. Thus, we use data pruning, i.e., selecting some solutions with better scores for training. The corresponding experimental validation will be presented in Section~\ref{sec:exp-additional-results}.

\textbf{Search algorithm.} After training the surrogate model, various methods can be used to obtain the final solution set. We default to using the popular NSGA-II~\cite{nsgaii} as the search algorithm. Additionally, we also consider employing other MOO algorithms, such as MOEA/D~\cite{moead}, NSGA-III~\cite{deb2013evolutionary}, and MOBO~\cite{qnehvi}.

\subsection{GP-Based Offline MOO Method}~\label{sec:5.2}

GP-based methods, often used in MOBO, are also promising for solving offline MOO problems. However, GP has a much higher computational complexity compared to DNN. Specifically, the computational complexity of learning a GP model is $\mathcal O(N^3+N^2D)$~\cite{gpml}, where $N$ is the number of data points and $D$ is the dimension of search space. Thus, directly using GP to offline MOO is not realistic, and data pruning is required. Similar to the data pruning approach in DNN-based methods, we use the non-dominated sorting in NSGA-II~\cite{nsgaii} to select $K$ data points at the front layers for learning. We will examine the impact of hyper-parameter $K$ on the performance in Figure~\ref{fig:mobo-ablate} of Appendix~\ref{app:results}. 
\fix{We use three mainstream MOBO frameworks for comparison:
1) Hypervolume-based method $q$NEHVI~\cite{qnehvi} selects solutions that can maximize the expected improvement in hypervolume, which is our default MOBO.
2) Scalarization-based method $q$ParEGO~\cite{qparego} randomly samples $q$ weight vectors to scalarize the objectives into $q$ single-objective problems and uses expected improvement to select points within each single-objective problem.
3) Information-theoretic-based method JES~\cite{jes} considers the information gain that maximally reduces the uncertainty in both the input and output spaces. 
On the special discrete tasks, we use Kendall kernel~\cite{bops} and transformed overlap kernel~\cite{antBO} as the kernel function of GP for permutation and sequence space, respectively.
MOBO employs NSGA-II~\cite{nsgaii} to generate a batch of solutions. MOBO-$q$ParEGO uses a single-objective genetic algorithm with specific operators to optimize the $q$ single-objective problem. MOBO-JES requires a stationary kernel, therefore Kendall kernel~\cite{bops} and transformed overlap kernel~\cite{antBO} cannot be utilized. Details are provided in Appendix~\ref{sec:moo-setting}.}

\section{Experiment}

In this section, we empirically examine the performance of different methods on our benchmark. We first introduce the experimental settings, and then report the main results on all the tasks. We also conduct additional experiments to show the challenges of offline MOO, compare training curves, study the effectiveness of data pruning, analyze the volume of data needed for MOBO, and investigate the influence of the search algorithm.

\begin{table*}[t!]
\caption{Average rank of different offline MOO methods on each type of task in Off-MOO-Bench, where the best and runner-up ranks are \textbf{bolded} and \underline{underlined}, respectively. Note that $\mathcal{D}\rm{(best)}$ denotes the best set in the training dataset, and the last column reports the average rank of each method on all the tasks.}\vspace{0.3em}
\centering
\resizebox{\linewidth}{!}{
\begin{tabular}{c|cccccc|c}
\toprule
Methods                & Synthetic       & MO-NAS          & MORL            & MOCO            & Sci-Design      & RE              & Average Rank     \\ \midrule
$\mathcal{D}$(best) & 12.17 $\pm$ 0.27 & 12.11 $\pm$ 0.05 & 9.00 $\pm$ 0.50 & \textbf{2.00 $\pm$ 0.14} & 8.38 $\pm$ 0.38 & 13.13 $\pm$ 0.07 & 10.03 $\pm$ 0.07 \\
End-to-End & 6.91 $\pm$ 0.03 & 8.37 $\pm$ 0.05 & 7.50 $\pm$ 2.00 & 6.75 $\pm$ 0.46 & 6.75 $\pm$ 1.12 & 7.50 $\pm$ 0.57 & 7.32 $\pm$ 0.01 \\
End-to-End + GradNorm & 8.25 $\pm$ 0.56 & 7.71 $\pm$ 0.08 & \underline{4.50 $\pm$ 1.00} & 7.61 $\pm$ 0.18 & 8.62 $\pm$ 0.50 & 10.53 $\pm$ 0.07 & 8.34 $\pm$ 0.01 \\
End-to-End + PcGrad & 7.88 $\pm$ 0.06 & 7.18 $\pm$ 0.39 & 10.50 $\pm$ 1.50 & 6.07 $\pm$ 0.64 & 8.69 $\pm$ 2.69 & 8.23 $\pm$ 0.17 & 7.51 $\pm$ 0.14 \\
Multi-Head & 6.38 $\pm$ 0.50 & 5.37 $\pm$ 0.37 & 6.25 $\pm$ 2.25 & 8.29 $\pm$ 0.21 & 9.19 $\pm$ 0.44 & 8.33 $\pm$ 0.40 & 7.00 $\pm$ 0.38 \\
Multi-Head + GradNorm & 7.78 $\pm$ 0.53 & 10.20 $\pm$ 0.04 & 11.00 $\pm$ 3.00 & 9.98 $\pm$ 0.30 & 9.06 $\pm$ 1.19 & 10.63 $\pm$ 0.17 & 9.63 $\pm$ 0.04 \\
Multi-Head + PcGrad & 8.61 $\pm$ 0.14 & 6.92 $\pm$ 0.55 & 10.50 $\pm$ 3.50 & 8.21 $\pm$ 0.36 & 9.38 $\pm$ 0.50 & 8.50 $\pm$ 0.17 & 8.09 $\pm$ 0.20 \\
Multiple Models & \textbf{4.05 $\pm$ 0.11} & \underline{4.93 $\pm$ 0.28} & 9.75 $\pm$ 0.75 & 6.34 $\pm$ 0.27 & \underline{5.62 $\pm$ 0.75} & \underline{4.50 $\pm$ 0.10} & \underline{5.02 $\pm$ 0.03} \\
Multiple Models + COMs & 9.81 $\pm$ 0.31 & 5.92 $\pm$ 0.34 & 7.00 $\pm$ 2.00 & 6.36 $\pm$ 0.50 & 8.38 $\pm$ 2.00 & 10.50 $\pm$ 0.50 & 8.09 $\pm$ 0.32 \\
Multiple Models + RoMA & 8.95 $\pm$ 0.05 & 5.00 $\pm$ 0.00 & 4.75 $\pm$ 2.25 & 8.14 $\pm$ 0.21 & 8.00 $\pm$ 1.38 & 6.30 $\pm$ 0.10 & 7.07 $\pm$ 0.02 \\
Multiple Models + IOM & \underline{6.11 $\pm$ 0.36} & \textbf{4.34 $\pm$ 0.34} & \textbf{3.75 $\pm$ 2.75} & 4.25 $\pm$ 0.04 & 7.19 $\pm$ 0.44 & \textbf{3.23 $\pm$ 0.03} & \textbf{4.61 $\pm$ 0.05} \\
Multiple Models + ICT & 9.11 $\pm$ 0.27 & 11.92 $\pm$ 0.29 & 4.75 $\pm$ 0.25 & 9.89 $\pm$ 0.46 & 8.62 $\pm$ 0.75 & 8.43 $\pm$ 0.30 & 9.64 $\pm$ 0.11 \\
Multiple Models + Tri-Mentoring & 7.83 $\pm$ 0.05 & 11.37 $\pm$ 0.47 & 5.25 $\pm$ 2.75 & 9.50 $\pm$ 0.00 & 9.38 $\pm$ 1.00 & 6.73 $\pm$ 0.20 & 8.77 $\pm$ 0.21 \\
MOBO & 9.09 $\pm$ 0.47 & 7.18 $\pm$ 0.55 & 10.50 $\pm$ 0.00 & 13.69 $\pm$ 0.08 & \textbf{5.44 $\pm$ 0.56} & 6.11 $\pm$ 0.29 & 8.64 $\pm$ 0.37 \\
MOBO-$q$ParEGO & 10.27 $\pm$ 0.23 & 11.47 $\pm$ 0.32 & N/A & 13.62 $\pm$ 0.04 & 9.44 $\pm$ 0.44 & 12.71 $\pm$ 0.33 & 11.68 $\pm$ 0.20 \\
MOBO-JES & 12.48 $\pm$ 0.05 & 16.00 $\pm$ 0.00 & N/A & \underline{3.00 $\pm$ 0.00} & 7.50 $\pm$ 6.50 & 8.04 $\pm$ 0.37 & 10.30 $\pm$ 0.44 \\ \bottomrule
\end{tabular}}\label{tab:main}
\end{table*}

\subsection{Experimental Settings}

\fix{The compared methods are introduced as follows. For NN-based methods, we consider the three types of models discussed in Section~\ref{sec:5.1}. 1) End-to-End models, including End-to-End, End-to-End + GradNorm~\cite{gradnorm}, and End-to-End + PcGrad~\cite{gradient-surgery}. 2) Multi-Head models, including Multi-Head, Multi-Head + GradNorm, and Multi-Head + PcGrad. 3) Multiple models, including Multiple Models, Multiple Models + COMs~\cite{coms}, Multiple Models + RoMA~\cite{ROMA}, Multiple Models + IOM~\cite{iom}, Multiple Models + ICT~\cite{ict}, and Multiple Models + Tri-Mentoring~\cite{tri-mentoring}. 
For GP-based methods, we use the three main types, including MOBO (i.e., $q$NEHVI~\cite{qnehvi}), MOBO-$q$ParEGO~\cite{parego}, and MOBO-JES~\cite{jes}.} 
All the advanced methods use data pruning by default.
After training the model, a search algorithm (which is NSGA-II~\cite{nsgaii} by default) is run in the model to generate a set of 256 solutions, which are then conducted by one batched evaluation and used to calculate the HV value \fix{(i.e., 100th percentile evaluations). 
The results of other settings, including 256 solutions with 50th percentile evaluations and 32 solutions with 100th percentile evaluations are provided in Appendix~\ref{app:results}.}
The model architecture and hyperparameters are consistently maintained across all tasks. Different operators are used for different search spaces, while the operator remains the same within the same search space across all the methods. We report the mean performance and standard deviation over five identical seeds (1000, 2000, ..., 5000) for all algorithms on all the tasks. \fix{Note that not all methods can be applied to every task in Off-MOO-Bench due to the long running time and high computational resource cost (for example, running out of GPU memory due to the high complexity), and we indicate this with ``N/A".} Detailed experimental settings are provided in Appendix~\ref{app:settings}.

\subsection{Main Results}

Table~\ref{tab:main} shows the average rank of all the compared methods on each type of task. Note that $\mathcal{D}\rm{(best)}$ denotes the best solution set (having the largest HV value) in the training set. Based on the average rank of all tasks (i.e., the last column of the table), we can observe that \fix{all the NN-based} offline MOO methods outperform the best solution set in the training set, i.e., $\mathcal{D}\rm{(best)}$, showcasing the feasibility and effectiveness of offline MOO. \fix{Multiple Models + IOM is the generally the best method (i.e., winner of Tables~\ref{tab:main} and~\ref{tab:avg-rank-32-100}, runner-up of Table~\ref{tab:avg-rank-256-50}), demonstrating the effectiveness of advanced offline SOO techniques}. 
However, optimizing in a discrete space can indeed be challenging, e.g., no method can achieve a better rank than $\mathcal{D}\rm{(best)}$ on MOCO, whose search space is a permutation space. This is primarily due to the complexity of modeling the discrete space in the surrogate model, which is also a significant challenge in offline SOO~\cite{bootgen}. 
\fix{Although GP-based methods can be applied to offline MOO by setting the number of iterations to 1 with a large batch size, their performance remains unsatisfactory. MOBO achieves only an average ranking of 8.64, while MOBO-$q$ParEGO and MOBO-JES perform even worse than $\mathcal{D}\rm{(best)}$. Targeted MOBO algorithms for offline MOO, such as selecting training data more effectively based on the statistics of the offline dataset and designing improved kernel functions, can be proposed based on our benchmark, which represents an interesting direction for future research.}
We can also find that no single method demonstrates a significant advantage, and even the best-performing method only has an average rank of 4.61. 
These findings indicate that there is still an ongoing challenge to further enhance the effectiveness of offline MOO.

\subsection{Additional Results}\label{sec:exp-additional-results}
In this section, we mainly aim to answer the question: What matters to the performance of offline MOO methods? Other results, including the analysis of data pruning, and the analysis on the influence of number of initial points of MOBO and the search algorithms, are provided in Appendix~\ref{app:results} due to space limitation.

\begin{figure}[t!]\centering
\includegraphics[width=0.8\linewidth]{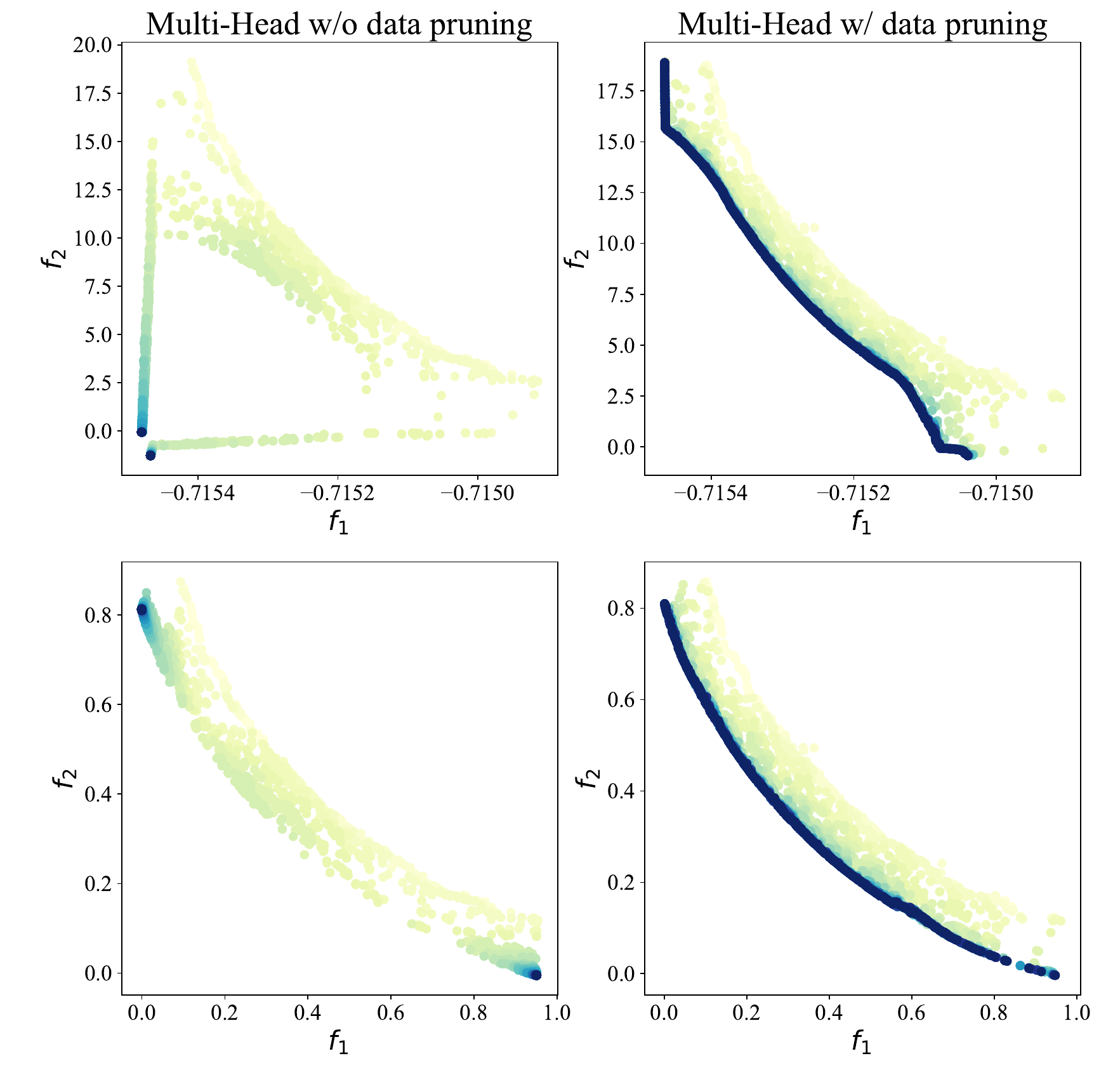}
\vspace{-1em}
\caption{Objective space visualization of Multi-Head model without (left column) and with (right column) data pruning on RE21, where the upper and bottom rows correspond to the surrogate objective space and real objective space, respectively. Each point denotes a solution in the search history, whose color gradually changes from yellow to blue based on the iteration rounds of the search algorithm.}\label{fig:challenges-MOO}
\vspace{-1em}
\end{figure}

\textbf{A key challenge of offline MOO} is that an inaccurate surrogate model will destroy the final performance. The output of offline MOO is a set of solutions that are non-dominated to each other. If the surrogate model is inaccurate, the Pareto-dominance relationship will be largely influenced. For example, if the model wrongly predicts that one solution is very good, then the solution will dominate all the other solutions, resulting in only few solutions in the final solution set and an extremely low HV value. In our experiments, we have found that the main reason of inaccurate surrogate model lies in the poor performance of learning those solutions with better objective values, i.e., elites. 
\fix{To address this issue}, we use data pruning to remove the solutions with worse objective values, allowing the model to focus more on learning from good regions and then obtaining a more accurate model. This will lead to a better final performance, as shown in Figure~\ref{fig:challenges-MOO}. The left and right columns denote the Multi-Head model without and with data pruning on the task of RE21, respectively. The upper-row shows the search process in the objective space of the surrogate model (i.e., proxy objective space), and the bottom-row shows their mapping in the real oracle objective space. We can observe that the model without data pruning has a phenomenon we discussed before, i.e., there are only two solutions in the final solution set. The model with data pruning performs much better, but still exhibits a certain degree of over-estimation which is also quite common in offline SOO~\cite{coms}. 
Thus, finding ways to mitigate such phenomenon is an important future direction in offline MOO. 
\fix{Detailed experiments and discussions about model collapse and data pruning are provided in Appendix~\ref{app:additional results}.}

\begin{figure}[t!]\centering
\includegraphics[width=0.49\linewidth]{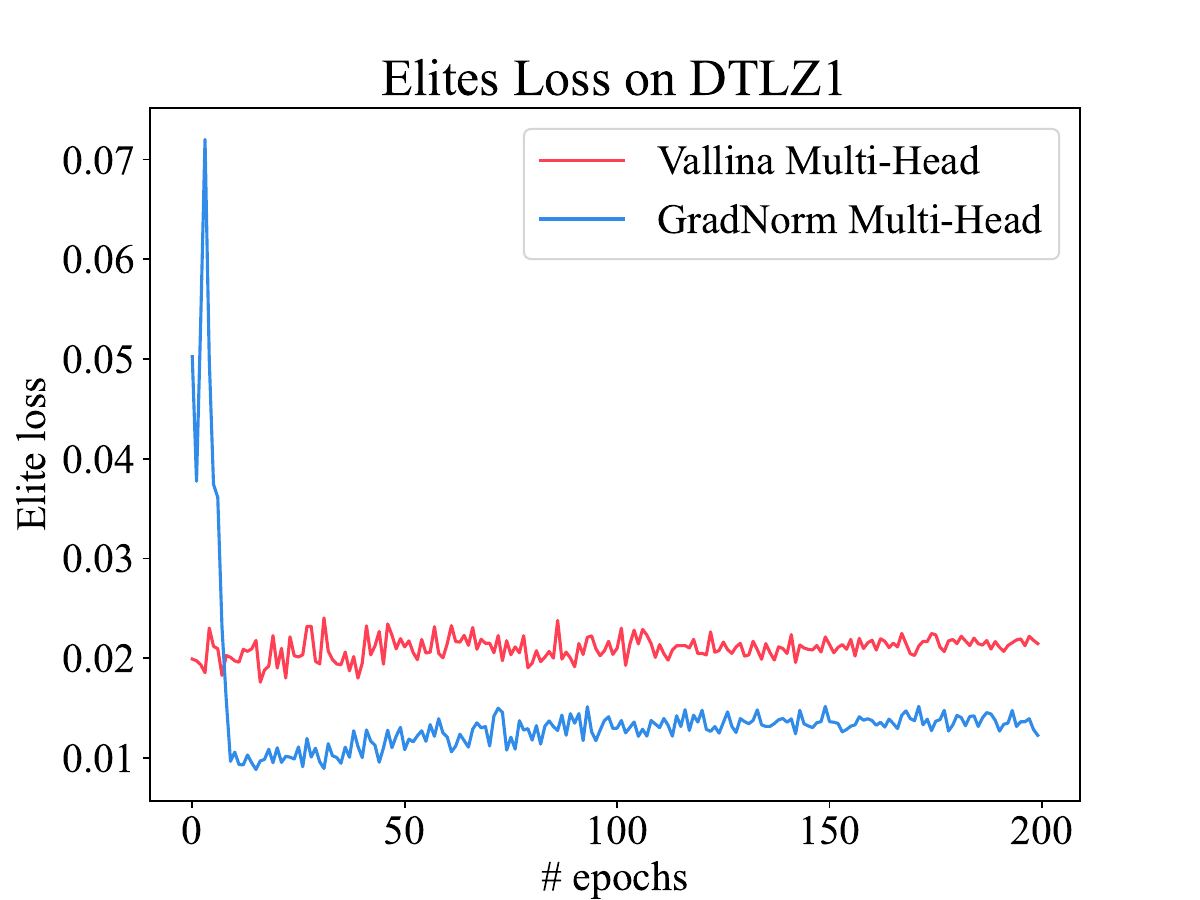}
\includegraphics[width=0.49\linewidth]{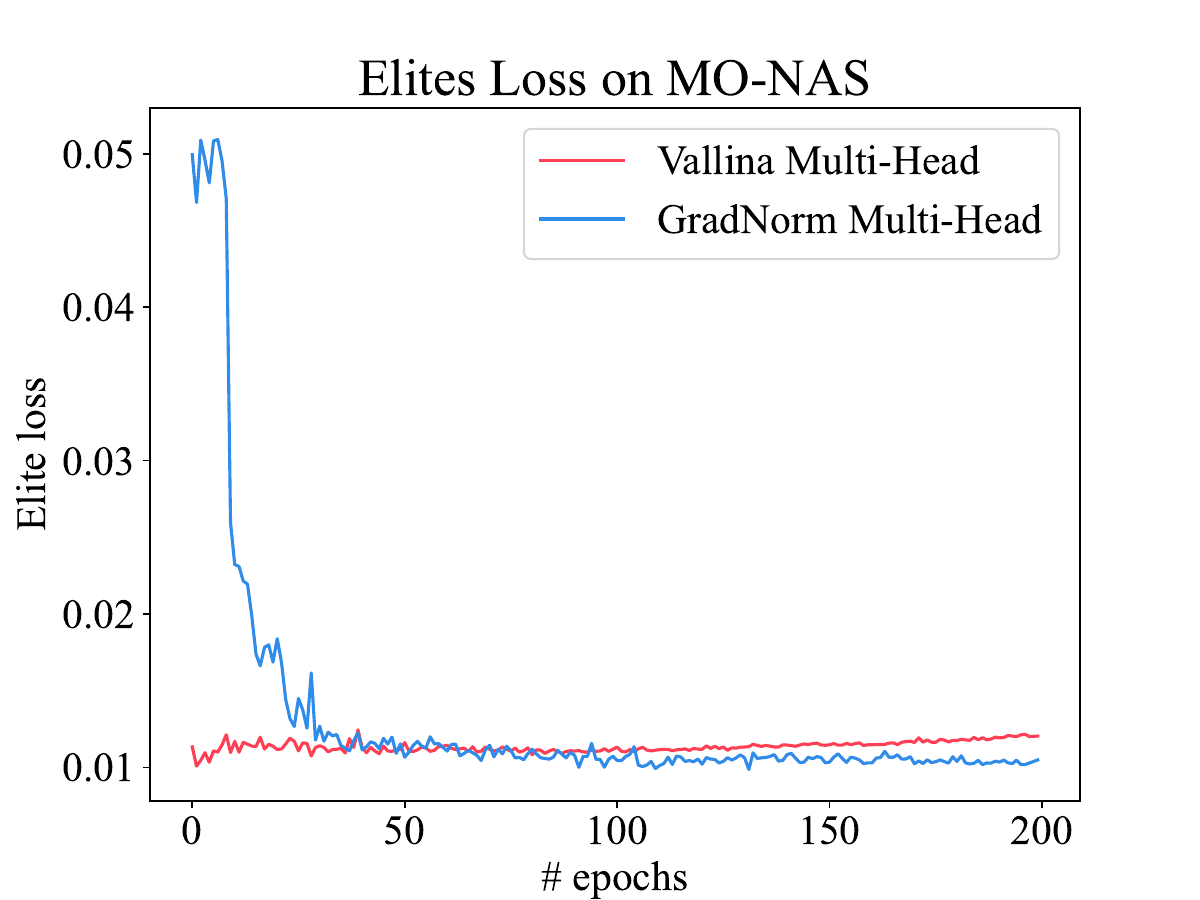} \\
\includegraphics[width=0.49\linewidth]{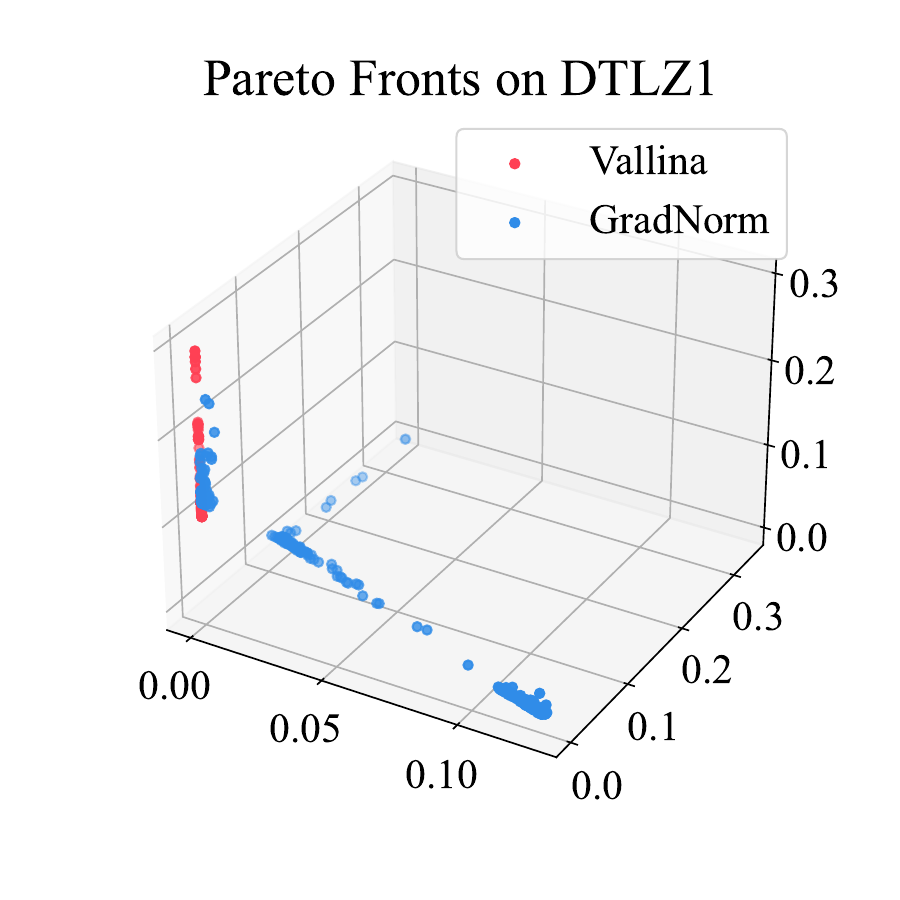}
\includegraphics[width=0.49\linewidth]{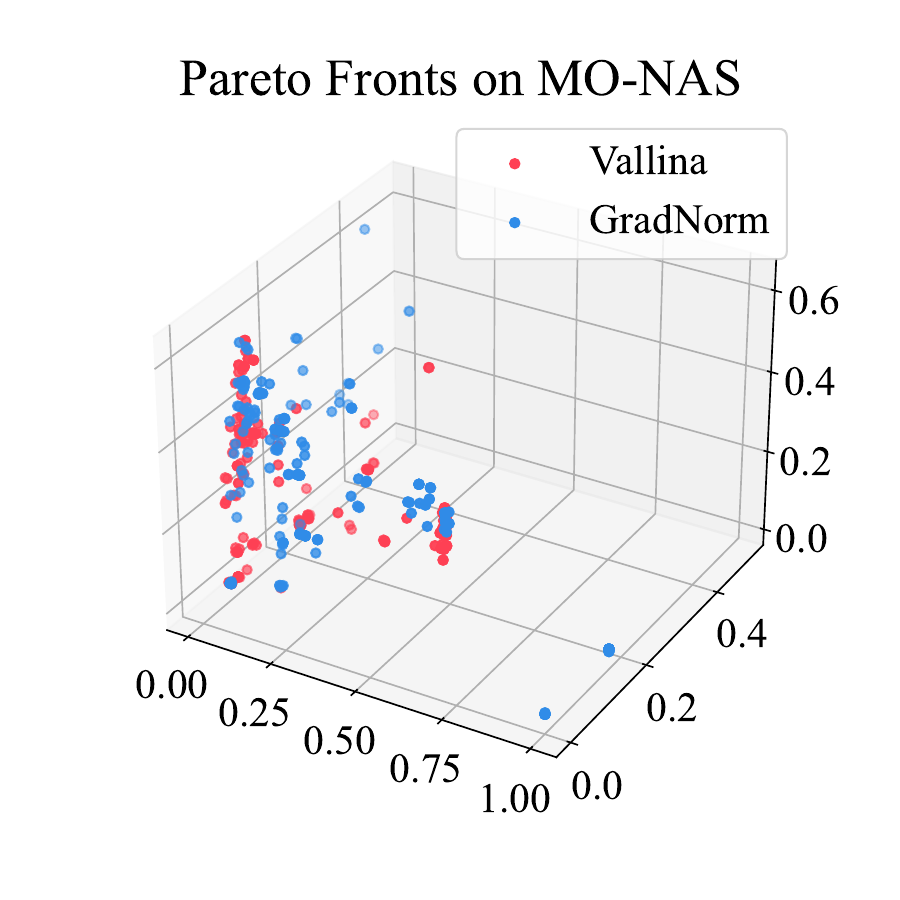}\vspace{-2em}
\caption{Elites loss changes (upper) and objective space visualizations of the final solution set (bottom), for Vanilla Multi-head model and Multi-head model with GradNorm on the two tasks, DTLZ1 and \fix{NAS-Bench-201-Test.}}\label{fig:training-loss}\vspace{-1.3em}
\end{figure}

\textbf{Learning curves.} Based on the above analysis, we have found that the prediction quality of elites has a significant impact on the final performance. To verify this, we compare the vanilla Multi-Head model with the Multi-Head model with GradNorm on two tasks, namely DTLZ1 (from synthetic function) and MO-NAS. Figure~\ref{fig:training-loss} shows the changes of the elites loss during the training phase and the visualization of the final solution set in the objective space. It can be clearly observed from the upper subfigures that GradNorm achieves smaller elites loss than vanilla Multi-Head. As a result, the solution set obtained by GradNorm has a generally better distribution, as shown in the bottom subfigures, and also has a better HV value. 
% On DTLZ-1, GradNorm and Vanilla have HV value 10.35 and 10.16, respectively; on MO-NAS, their HV values are 10.65 and 10.13, respectively.  

\section{Discussion}

\textbf{Conclusion.} In this paper, we emphasize the significance of offline MOO and provide the first benchmark that encompasses a range of crucial offline MOO tasks, from synthetic functions to real-world applications. Additionally, we introduce a framework of offline MOO methods and analyze the different components. Extensive experiments validate the efficacy of these methods. 
In the future, we will incorporate \fix{more analysis (e.g., the influence of reference points~\cite{refpoint,refpoint2})}, more demanding tasks (e.g., DNA sequence designs and industrial applications), more learning-based MOO algorithms (e.g., Pareto set learning~\cite{psl-1,psl-3} and MO-GFlowNets~\cite{MOGFN}), \fix{and more advanced offline SOO algorithms (e.g., LEO~\cite{leo}, and BRAID~\cite{uehara2024bridging})} into our benchmark.

\textbf{Future works of offline MOO.} Based on our experimental results and analyses, there are many worthwhile directions for future exploration. Here we discuss some challenges of offline MOO and hope to shed some light on future works.
\begin{enumerate} 
    \item \textbf{Mixed search space.} Most search spaces of the current problems are either continuous or discrete. However, in practice, many problems involve mixed variables~\cite{mixed}, which pose significant challenges for offline MOO, especially in constructing accurate surrogate models.
    \item \textbf{Large-scale (high-dimensional) optimization.} The high-dimensionality of the search space is a common challenge of black-box optimization~\cite{high-dim}. Our experimental results indicate that the current offline MOO methods do not perform well on large-scale problems, e.g., no method surpasses the best value of the training set on MOCO. Exploring effective techniques such as dimensionality reduction~\cite{rembo,mctsvs} to efficiently solve large-scale problems is an important future direction.
    \item \textbf{Constrained optimization.} Many real-world MOO tasks come with strict constraints~\cite{constrained}, making surrogate model learning and search challenging. Our current approach is rather simplistic, which directly discards solutions that do not satisfy the constraints. Employing more efficient constraint handling strategies would significantly improve the performance.
    \item \textbf{Noisy optimization.} The black-box evaluations of numerous real-world problems involve intricate processes, which often suffer from inaccuracies due to the inevitable presence of noise~\cite{noisy,noisy-ea}. The noise may have a detrimental effect on the quality of the offline dataset, presenting a significant challenge that needs to be addressed.
    \item \textbf{Few-shot optimization.} Some application scenarios do not have strict limitations on the number of evaluations but allow for a few batches~\cite{few-shot}. How to utilize a limited number of iterative evaluations to adapt the surrogate model is indeed a crucial task for future work.
\end{enumerate}

\section*{Impact Statement}
This paper presents work whose goal is to advance the field of multi-objective optimization and machine learning. There are many potential societal consequences of our work, none of which we feel must be specifically highlighted here.

\section*{Acknowledgements}
\fix{The authors want to thank the anonymous reviewers for their helpful comments and suggestions. This work was supported by the National Science and Technology Major Project (2022ZD0116600) and National Science Foundation of China (62276124).}

\bibliography{main}
\bibliographystyle{icml2024}

\clearpage
\clearpage
\newpage
\appendix
\onecolumn
\section{Detailed settings}\label{app:settings}
In this section, we provide detailed settings of Off-MOO-Bench regarding data collection, training set formulation, and default settings of offline MOO methods.

\subsection{Dataset Collection} \label{sec:data-collect-detail}
As discussed in Section~\ref{sec:data-collect}, only using the expert MOEAs, i.e., NSGA-II, MOEA/D, and NSGA-III, may result in obtaining solutions with good quality, which generates a significant difference between data distribution and diverse reality distribution. Inspired by~\cite{d4morl}, we propose an amateur survival operator for MOEAs. We first use the expert MOEAs to perform generic mating and generate offspring. Assume that the current population has $\mu$ individuals and the offspring population has $k$ individuals. With a given probability $p$, we choose the $\mu$ among $(\mu + k)$ individuals according to non-dominated sorting of NSGA-II~\cite{nsgaii} to form the next population. 
Otherwise, with probability $1-p$, we survive the $\mu$ best-non-dominated individuals, as the Survival operator in NSGA-II. 
After a small amount of number of generations (e.g. 1 or 5), we collect the current population to form the final dataset. Typically, we use NSGA-II with the amateur survival operator as the amateur collection algorithm for all tasks of synthetic functions, MOCO, scientific design, and most tasks of RE, except for RE34, where we use NSGA-III due to its better performance in obtaining dataset with diversity.

For \fix{NAS-Bench 201~\cite{nas201} and NARTS~\cite{narts} in MO-NAS}, since the size of their search space is \fix{limited (i.e., 15625 for NAS-Bench 201 and 32678 for NARTS)}, we directly iterate the whole search space and collect query-answers of MO-NAS-Bench~\cite{mo-nas}. 
For MORL tasks, since our goal is to learn directly from policies to rewards, the algorithm for data collection proposed by D4MORL~\cite{d4morl} is not suitable for us. Thus, we run the SOTA MORL algorithm PGMORL~\cite{pgmorl} with 100 different seeds and collect the policies. 
For the tasks of scientific design, we use the Amateur-NSGA-II (as discussed above) to collect part of the dataset, and randomly sample over the whole search space to form the other part. 

\subsection{Training Set Construction}
In realistic scientific and industrial scenarios, we usually hope to use offline collected dataset to obtain better designs than offline ones. Thus, similar to ~\cite{design-bench}, we remove the top solutions sorted by NSGA-II ranking with a given percentile $K$, where $K$ varies according to different tasks and is usually set 40\%, except for Molecule with 1.2\%, RFP and Regex with 20\%, and MO-CVRP with 55\%. Besides, we perform normalization within each objective of each problem because different objectives can have different scales, which may result in imbalanced model update.

\subsection{Training Details}

For NN-based model, similar to Design-Bench~\cite{design-bench}, the End-to-End network structure is:
\begin{align*}
        \text{input} &\to \textbf{MLP}(2048)\to \textbf{relu}
        \to \textbf{MLP}(2048)\to \textbf{relu} \to
        \textbf{MLP}(\text{number of objectives}).
\end{align*}

The Multi-Head model is constructed by two parts of neural networks, feature extractor and task head. For feature extractor, the structure is:
\begin{align*}
        \text{input} &\to \textbf{MLP}(2048)\to \textbf{relu}
        \to \textbf{MLP}(2048).
\end{align*}
For task head, the structure is:
\begin{align*}
        \text{features with 2048 dimensions} &\to \textbf{relu}
        \to \textbf{MLP}(1).
\end{align*}
The network structure of multiple models is 
\begin{align*}
        \text{input} &\to \textbf{MLP}(2048)\to \textbf{relu}
        \to \textbf{MLP}(2048)\to \textbf{relu} \to
        \textbf{MLP}(\text{1}).
\end{align*}
We use MSE as loss function and optimize by Adam with learning rate $\eta=0.001$ and learning-rate decay $\gamma=0.98$. The DNN model is trained w.r.t. offline dataset for 200 epochs with a batch size of 32.

For the End-to-End + GradNorm method, we perform gradient normalization on the last MLP layer. For the Multi-Head + GradNorm method, we perform gradient normalization on the last MLP layer of the feature extractor, as in MTL~\cite{gradnorm}.

For GP-based methods, we choose the top 100 solutions by NSGA-II ranking to initialize the GP model. Since the performance of a GP model is sensitive to the number of initialized points, we conduct ablation studies in Figure~\ref{fig:mobo-ablate}.

Among all tasks in Off-MOO-Bench, Molecule and MOCO have constraints. For Molecule, since we optimize in the latent space following~\citet{lamoo}, we cannot judge if a solution is feasible in the latent space. Thus, we first obtain a batch of 256 solutions generated by the algorithm and then filter out the infeasible ones during evaluation. For MOCO tasks, since we use the Start-From-Zero repair operator, the constraint is avoided.

\subsection{MOO Settings} \label{sec:moo-setting}
\fix{For DNN-based surrogate models, after the model is trained, we use multi-objective evolutionary algorithms (MOEAs) to optimize inside the trained model. To obtain $K$ (approximately) Pareto-optimal solutions, we set the size of population to $K$ and initialize the population with $K$ non-dominated solutions in the offline dataset, and the algorithm searches for 50 generations. We use different genetic operators for different types of tasks.}
Specifically, for continuous tasks (i.e., synthetic functions, RE, NAS-Bench-201-Test, MO-Portfolio, MORL, and Molecule), we use the default genetic operators of NSGA-II implemented in PyMOO~\cite{pymoo}, \fix{i.e., Simulated Binary Crossover (SBX) and Polynomial Mutation (PM)}.
For discrete tasks in MOCO (i.e., MO-TSP, MO-CVRP, and MO-KP), since the search space is combinatorial, where each solution in the three problems can be represented as a permutation, we use Order-Crossover as the crossover operator, Inversion-Mutation as the mutation operator, and for MO-TSP and MO-CVRP problems, we utilize the Start-From-Zero repair operator to make sure that the salesman starts from the deposit. 
\fix{For C-10/MOP and IN-1K/MOP test suites in MO-NAS, we use the suggested genetic operators from the source code of~\citet{mo-nas}, i.e., PM for integer with $\eta=20$ and SBX for integer with $\eta=30$.}
For Regex\fix{, ZINC,} and RFP tasks, we use the local mutation operator implemented by LaMBO~\cite{lambo}, and SBX for integer as in~\citet{lambo}. 

\fix{For GP-based surrogate models, we use different methods to optimize the acquisition function for different types of tasks. Specifically, for continuous tasks (i.e., synthetic functions, RE, NAS-Bench-201-Test, MO-Portfolio, MORL, and Molecule), we use gradient-based methods (i.e., L-BFGS-B~\cite{l-bfgs-b}) to optimize the acquisition function, which is the default acquisition function optimization method implemented in BoTorch~\cite{botorch}. For discrete tasks, we use MOEAs to optimize the acquisition function. Our default MOBO employs NSGA-II~\cite{nsgaii} to generate a batch of solutions that minimize the lower confidence bound of GP, where we set the size of population to $K$ and initialize the population with $K$ non-dominated solutions in the offline dataset, and the algorithm searches for 500 generations with SBX crossover and PM mutation to obtain the final solutions. For MOBO-$q$ParEGO, we use single-objective evolutionary algorithms to optimize the acquisition function. Specifically, we first initialize the population with 50 randomly sampled points, and then search for 500 generations to obtain the best solution for each scalarized single-objective problem. 
The genetic operators for discrete spaces are as same as the ones in evolutionary search algorithms inside DNN-based surrogate models. Note that MOBO-JES cannot run in discrete tasks, since it requires a stationary kernel and thus Kendall kernel and transformed overlap kernels cannot be utilized.}

The implementations of NSGA-II, MOEA/D, and NSGA-III are from the open-source repository PyMOO~\cite{pymoo}. 
The implementation of MOBO is inherited from BoTorch~\cite{botorch}.

\section{Detailed Tasks}\label{app:task}
In this section, we provide details of different MOO tasks adopted in our experiments. Notably, certain maximization tasks undergo transformation into minimization problems through the multiplication of $-1$.  The reference points $\bm{r}$ for majority of tasks are set in such a way that $(r_i - z^{i}_{\min})/(z^{i}_{\max}-z^{i}_{\min})=1.1$ , except that for MORL tasks, $(r_i - z^{i}_{\min})/(z^{i}_{\max}-z^{i}_{\min})=2.0$, where $r_i$ denotes the value on the $i$-th dimension of the reference point $\bm{r}$, and $z^{i}_{\max}$ and $z^{i}_{\min}$ are the maximum value and minimum value of the $i$-th objective in the collected data, respectively. It means that after normalization, the reference point becomes $(1.1,\dots,1.1)$ or $(2.0,\dots,2.0)$.

\subsection{Synthetic Function}
Various widely-used synthetic functions in MOO literature are employed to evaluate the algorithms. Specifically, the following benchmark problems are used: \fix{DTLZ1-7~\cite{dtlz}, ZDT1-4, ZDT6~\cite{zdt} , Omni-test~\cite{omnitest} and VLMOP1-3~\cite{vlmop}}. The solution spaces for all synthetic problems are continuous. The detailed problem information, Pareto front shape and reference point can be found in Table~\ref{table:synthetic}. Note that the concave (2d) Pareto front for DTLZ5 and DTLZ6 indicates that  the Pareto front takes the form of a degenerated 2-dimensional curve within a 3-dimensional objective space.

\begin{table}[htbp]
\centering
\caption{Problem information and reference point for synthetic functions. }
\begin{tabular}{llllll}
\toprule
Name     & $D$ & $m$ & Type & Pareto Front Shape & Reference Point \\ \midrule
DTLZ1    & 7   & 3 &Continuous  & Linear       &    (558.21, 552.30, 568.36)             \\
DTLZ2    & 10  & 3   &Continuous & Concave      &    (2.77, 2.78, 2.93)             \\
DTLZ3    & 10  & 3    &Continuous & Concave      &       (1703.72, 1605.54, 1670.48)          \\
DTLZ4    & 10  & 3  &Continuous  & Concave      &         (3.03, 2.83, 2.78)        \\
DTLZ5    & 10  & 3  &Continuous  & Concave (2d)  &        (2.65, 2.61, 2.70)         \\
DTLZ6    & 10  & 3   &Continuous & Concave (2d)  &         (9.80, 9.78, 9.78)        \\
DTLZ7    & 10  & 3   &Continuous & Disconnected &          (1.10, 1.10, 33.43)       \\
ZDT1     & 30  & 2   &Continuous & Convex       &         (1.10, 8.58)        \\
ZDT2     & 30  & 2   &Continuous & Concave      &          (1.10, 9.59)       \\
ZDT3     & 30  & 2   &Continuous & Disconnected &        (1.10, 8.74)         \\
ZDT4     & 10  & 2   &Continuous & Convex       &          (1.10, 300.42)       \\
ZDT6     & 10  & 2   &Continuous & Concave      &         (1.07, 10.27)        \\
Omnitest & 2   & 2   &Continuous & Convex       &         (2.40, 2.40)        \\
%Kursawe  & 3   & 2   &Continuous & Disconnected &         (-3.39, 28.39)        \\
VLMOP1   & 1   & 2 & Continuous & Concave      & (4.0, 4.0) \\
VLMOP2   & 6   & 2 &Continuous  & Concave      &        (1.10, 1.10)         \\
VLMOP3   & 2   & 3  &Continuous & Disconnected &   (9.07, 66.62, 0.23)   \\  \bottomrule
\end{tabular}\label{table:synthetic}
\end{table}

\subsection{MO-NAS}\label{sec:mo-nas}
MO-NAS~\cite{mo-nas} automates the exploration of optimal neural network architectures to enhance multiple model metrics for specific tasks. In our experiments, we conduct \fix{a toy example, named NAS-Bench-201-Test,} to optimize three objectives: prediction error, number of parameters, and edge GPU latency, on the CIFAR-10 dataset~\cite{cifar10}. The prediction error metric primarily assesses the model's performance, the number of parameters gauges the model's scale, and the GPU latency is a hardware metric evaluating the efficiency of GPU during model execution. The search space is from~\citet{nas201}. Given a macro skeleton of the neural network and a directed acyclic graph structure for each cell, our objective is to explore the operations (edges) within the cell. Each cell contains 6 edges, and there are 5 predefined operation options for each edge (zeroize, skip-connect, $1\times 1$ convolution, $3\times3$ convolution, and $3\times 3$ average pool). Consequently, the search space is a 6-dimensional discrete space, where each dimension can take 5 values, resulting in a total of $5^6=15625$ possible solutions.
\fix{The data of NAS-Bench-201-Test}, corresponding error and number of parameters are sourced from~\citet{nas201}. Additionally, the edge GPU latency data is obtained from~\citet{hwnas}. The reference point is $\bm{r}=(98.48, 1.68, 12.81)$ .

\fix{Furthermore, we consider two test suites from~\citet{mo-nas}, i.e., C-10/MOP and IN-1K/MOP, which contain 18 tasks. The search spaces of these test suites vary from~\textit{micro} search spaces to \textit{macro} search spaces. Specifically, \textit{micro} search spaces are used to create a basic building block, often called a \textit{cell}, which is used repeatedly to build a full deep neural network (DNN) based on a set pattern; \textit{macro} search spaces are used to design the overall structure of the network, while the individual layers are designed using  well-established methods. \textit{Micro} search spaces include NAS-Bench-101~\cite{nas101}, NAS-Bench-201~\cite{nas201}, and DARTS~\cite{darts-nas}. \textit{Macro} search spaces include NATS~\cite{narts}, ResNet50~\cite{resnet50}, Transformer~\cite{autoformer}, and MNV3~\cite{resnet50}. Detailed information of these search spaces $\mathcal{X}$ can be found in Table~\ref{table:nas-search-space}. The detailed problem information and reference point of C-10/MOP1-9 and IN-1K/MOP1-9 tasks can be found in Table~\ref{table:nas-info}. 
Note that we transform the discrete search space of NAS-Bench-201-Test into a continuous logit space, which is the strategy in Design-Bench~\cite{design-bench} for handling discrete categorical tasks. However, it cannot be applied to all tasks in MO-NAS, since it requires that all dimensions have the same number of categories, while the number of categories in most tasks of MO-NAS differ from dimensions.}

\begin{table}[htbp]
\centering
\caption{An overview of the search spaces in MO-NAS tasks. }
\begin{tabular}{cccc}
\toprule
Search space $\mathcal{X}$     & Type & $D$ & $|\mathcal{X}|$ \\ \midrule
NAS-Bench-101    & micro   & 26 & 423K               \\
NAS-Bench-201    & micro   & 6 & 15.6K               \\
NATS    & macro   & 5 & 32.8K               \\
DARTS   & micro   & 32 & $\sim 10^{21}$ \\
ResNet50   & macro   & 25 & $\sim 10^{14}$ \\
Transformer   & macro   & 34 & $\sim 10^{14}$ \\
MNV3   & macro   & 21 & $\sim 10^{20}$ \\
\bottomrule
\end{tabular}\label{table:nas-search-space}
\end{table}

\begin{table}[htbp]
\centering
\caption{Problem information and reference point for C-10/MOP1-9 and IN-1K/MOP1-9 tasks.}
\resizebox{\linewidth}{!}{
\begin{tabular}{lllll}
\toprule
Problem     & Search space $\mathcal{X}$ & $D$ & $m$ & Reference Point \\ \midrule
C-10/MOP1    & NAS-Bench-101   & 26 & 2 & $(3.49\times 10^{-1}, 3.14\times 10^7)$        \\
C-10/MOP2    & NAS-Bench-101   & 26 & 3 & $(9.05\times 10^{-1}, 3.05\times 10^7, 8.97\times 10^9)$        \\
C-10/MOP3    & NATS   & 5 & 3 & $(2.31\times 10^{1}, 7.14\times 10^{-1}, 2.74\times 10^2)$        \\
C-10/MOP4    & NATS   & 5 & 4 & $(2.31\times 10^{1}, 7.14\times 10^{-1}, 2.74\times 10^2, 2.12\times 10^{-2})$        \\
C-10/MOP5    & NAS-Bench-201   & 6 & 5 & $(9.03\times 10^{1}, 1.53\times 10^{0}, 2.20\times 10^2, 1.17\times 10^{1}, 4.88\times 10^1)$       \\
C-10/MOP6    & NAS-Bench-201   & 6 & 6 & $(9.03\times 10^{1}, 1.53\times 10^{0}, 2.20\times 10^2,$ $1.05\times 10^1, 2.23\times 10^0, 2.76\times 10^1)$        \\
C-10/MOP7    & NAS-Bench-201   & 6 & 8 & $(9.03\times 10^{1}, 1.53\times 10^{0}, 2.20\times 10^2, 1.17\times 10^{1},$ \\
& & & &$4.88\times 10^1, 1.05\times 10^1, 2.23\times 10^0, 2.76\times 10^1)$       \\
C-10/MOP8    & DARTS   & 32 & 2 & $(2.61\times 10^{-1}, 1.55\times 10^6)$        \\
C-10/MOP9    & DARTS   & 32 & 3 & $(4.85\times 10^{-2}, 3.92\times 10^5)$        \\ \midrule
IN-1K/MOP1    & ResNet50   & 25 & 2 & $(2.81\times 10^{-1}, 3.95\times 10^7)$        \\
IN-1K/MOP2    & ResNet50   & 25 & 2 & $(2.80\times 10^{-1}, 1.15\times 10^{10})$        \\
IN-1K/MOP3    & ResNet50   & 25 & 3 & $(2.81\times 10^{-1}, 3.87\times 10^{7}, 1.26\times 10^{10})$        \\
IN-1K/MOP4    & Transformer   & 34 & 2 & $(1.83\times 10^{1}, 7.25\times 10^{7})$        \\
IN-1K/MOP5    & Transformer   & 34 & 2 & $(1.83\times 10^{1}, 1.49\times 10^{10})$       \\
IN-1K/MOP6    & Transformer   & 34 & 3 & $(1.83\times 10^{1}, 7.10\times 10^7, 1.48\times 10^{10})$        \\
IN-1K/MOP7    & MNV3 & 21 & 2 & $(2.64\times 10^{-1}, 9.98\times 10^{6})$ \\
IN-1K/MOP8    & MNV3 & 21 & 3 & $(2.65\times 10^{-1}, 1.00\times 10^7, 1.34\times 10^9)$        \\
IN-1K/MOP9    & MNV3 & 21 & 4 & $(2.65\times 10^{-1}, 1.03\times 10^7, 1.31\times 10^9, 6.30\times 10^1)$   \\
\bottomrule
\end{tabular}}
\label{table:nas-info}
\end{table}

\subsection{MORL}
MORL is an approach where the training of an agent focuses on simultaneously maximizing multiple cumulative rewards in some control environment. 
\fix{The primary purpose of proposing the MORL problem in our benchmark is to examine the performance of offline MOO in high-dimensional continuous spaces. Different from the D4MORL benchmark~\cite{d4morl}, we focus on direct policy parameter search, ignoring some properties of MDP. Note that using numerous neural network parameters as a search space for black-box optimization presents a significant optimization challenge, which is profoundly significant for offline MOO itself.} In our experiments, we consider two locomotion tasks namely MO-Swimmer and MO-Hopper, within the widely used MuJoCo benchmark~\cite{mujoco}. The search space consists of the parameters of the policy network for each environment as defined in~\cite{pgmorl}, whose dimension is much higher than other tasks.

\textbf{MO-Swimmer.}
This is a two-objective task  with an eight-dimensional state space and a two-dimensional action space. 
The two objectives are forward speed and energy efficiency, denoted as $\bm R = [R^s, R^e]$.
The search space is the 9734-dimensional policy network for MO-Swimmer. At time $t$, the agent is at position $(x_t, y_t)$ and takes an action $a_t$. Then, the instantaneous rewards at time $t$ are defined as:
\begin{align}
      R^s_t  = (x_t-x_{t-1})/0.05,
    \nonumber
    \\
      R^e_t = 0.3 - 0.15 \times \sum_{k} a_k^2.
      \nonumber
\end{align}

The reference point $\bm{r} = (267.67, 99.05)$ after multiplying $-1$.

\textbf{MO-Hopper.}
This is a two-objective task  with an eleven-dimensional state space and a three-dimensional action space. The two objectives are forward speed and jumping height, denoted as $\bm R = [R^s, R^j]$. The search space is the 10184-dimensional policy network for MO-Hopper. At time $t$, the agent is at position $(x_t, h_t)$ and takes an action $a_t$. Then, the instantaneous rewards at time $t$ are defined as:
\begin{align}
      R^s_t  = 1.5\times(x_t-x_{t-1})/0.01 + 1 - 2\times 10^{-4}\sum_{k}a_k^2,
    \nonumber
    \\
      R^j_t = 12\times (h_t- h_0)/0.01 + 1 - 2\times 10^{-4}\sum_{k}a_k^2,
      \nonumber
\end{align}
where $h_0=1.25$ is the initial height. The reference points $\bm{r} = (1489.01, 4734.48)$ after multiplying $-1$.

\subsection{MOCO}
We evaluate the algorithms on three typical \fix{discrete} MOCO problems, i.e., the multi-objective traveling salesman problem (MO-TSP)~\cite{motsp}, multi-objective capacitated vehicle routing problem (MO-CVRP)~\cite{mocvrp} and multi-objective knapsack problem (MO-KP)~\cite{mokp}, and one continuous MOCO problem, i.e., multi-objective portfolio problem (MO-Portfolio). The search spaces for the three discrete problems are formulated as permutation spaces, where the parameters of problem instance are randomly generated similar to~\cite{nhde}. \fix{Additional, for the MO-TSP problem, we also consider its tri-objective variant, as in~\cite{nhde}.}
\fix{The MO-Portfolio problem has a continuous search space, i.e., $[0,1]^n$, to represent the weights of portfolio allocation. Historical stock prices data of each portfolio is provided by~\citet{pymoo}.}

\textbf{MO-TSP} has \fix{$n=500,100,50,20$} nodes, and each node has two sets of two-dimensional coordinates, where the $i$-th objective value of the solution is calculated with respect to the $i$-th set of coordinates. The coordinates are generated uniformly from $[0, 1]^2$. Hence, this is a $n$-dimensional two-objective permutation optimization problem. The reference point is $\bm r=(255.18, 248.44)$.
    
\textbf{MO-CVRP} has \fix{$n=100,50,20$} customer nodes and a depot node, with each node featured by a two-dimensional coordinate and each customer node associated with a demand. Following the common practice, we consider two objectives, i.e., the total tour length and the longest length of the route. The coordinates and demands are generated uniformly from $[0, 1]^2$ and $\{0,\dots,9\}$, respectively. The capacity of vehicle is set to $50$. Each solution is represented as a $n$-dimensional permutation. 
For the evaluation of each solution (permutation), a vehicle departs from the depot and travels in the order specified by the permutation of customers. It accumulates customer capacity along the path, returning to the depot before reaching a customer in the permutation whose capacity exceeds the limit. The vehicle then continues from the depot, following the point after the last visited customer in the permutation. This process continues until completion. This is a $n$-dimensional two-objective permutation optimization problem. The reference point is $ \bm r=(49.19, 9.58)$.

\textbf{MO-KP} has \fix{$n=200,100,50$} items, with each taking a weight and two separate values. The $i$-th objective is to maximize the sum of the $i$-th values under the constraint of not exceeding the knapsack capacity. The weight and value of each item are generated uniformly from $[0,1]$. The capacity is set to $25$. Each solution is represented as a $200$-dimensional permutation. For the evaluation of each solution (permutation), we put the first $k$ items in the permutation into the knapsack, such that including the $(k+1)$-th item exceeds the knapsack capacity, while the first $k$ items remain within the capacity limit. This is a $n$-dimensional two-objective permutation optimization problem. The reference point is $ \bm r = (-7.85, -8.99)$ after multiplying $-1$.

\fix{\textbf{MO-Portfolio} has $n=20$ types of portfolios, with each taking an input as its corresponding weight. 
Here we consider the portfolio allocation problem based on the Markowitz Mean-Variance Portfolio Theory~\cite{portfolio} with two objectives, where the overall performance of a portfolio can be assessed through the expected return and overall risk of its assets. Geometrically, the expected return of a portfolio is defined as the average return of its assets and the risk is defined as the standard deviation. 
Additionally, in order to ensure that portfolio allocations are valid, we provide a repair operator that modifies the portfolio weights to ensure that they sum to 1 (as a common constraint in portfolio optimization) and no weights are smaller than a threshold $\theta=0.001$. The reference point is $\bm r = (0.29, -0.13)$.}

\subsection{Scientific Design}
\textbf{Molecule design.}
This is a two-objective molecular generation task~\cite{lamoo}. The task is to optimize the activity against biological targets GSK3$\beta$ and JNK3. The search space is a $32$-dimensional continuous latent space. The solutions in the latent space will be decoded into molecular strings and evaluated by a pre-trained decoder from~\citet{motifs}. 
The reference point $ \bm r=(0.09, 0.04)$.

\textbf{Protein design.}
We have incorporated two protein sequence design challenges outlined in~\cite{lambo}. The sequence optimization task starts with a base sequence pool $P$ of initial sequences, which are modified to produce new candidate
sequences. 
The optimization problem is restructured into the following nested decisions: 1) Choose a base sequence from the pool; 2) Choose which positions on the sequence to change; 3) Choose the operations to change the token at those positions; 4) If the operation is substitution or insertion, then select the tokens to substitute or insert.
Hence, the search space can be formalized as a four-dimensional space, encompassing choices for the base sequence, sequence positions, operations, and tokens.

For \textbf{Regex}, there are $16$ base sequences, $73$ sequence positions, $20$ types of tokens, and $3$ operations (substitution, deletion or insertion). So the search space has a size of $|\mathcal{X}|=16\times 72\times 20\times 3=69,120$. The goal is to maximize the counts of three predetermined bigrams. The reference point $\bm r=(1.11, 1.25, 1.21)$.

\fix{For \textbf{ZINC}, there are $16$ base sequences, $257$ sequence positions, $106$ types of tokens, and $3$ operations (substitution, deletion or insertion). So the search space has a size of $|\mathcal{X}|=16\times 257\times 106\times 3=1,307,616$. The goal is to maximize the  the octanol-water partition coefficient (logP) and QED (quantitative estimate of druglikeness). The reference point $\bm r=(1.36, 2.25)$.}

For \textbf{RFP}, there are $43$ base sequences, $489$ sequence positions, $20$ types of tokens, and $1$ operations (substitution). So the search space has a size of $|\mathcal{X}|=43\times 489\times 20\times 1=420,540$. The goal is to maximize the solvent-accessible surface area (SASA) and the stability of the RFP. The reference point $\bm r=(4.80, 4.54)$.

\subsection{RE}
We also conduct experiments on seven real-world multi-objective engineering design problems adopted from RE suite~\cite{RE}. These problems serve as practical application in various fields. The search spaces for the problems are continuous except for RE23, which has a mixed solution space (2 variables as integers and 2 as continuous values). The detailed problem information and reference points can be found in Table~\ref{table:re}.

\begin{table}[htbp]
\centering
\caption{Problem information and reference point for RE problems.}
\resizebox{\linewidth}{!}{
\begin{tabular}{lllllp{6cm}}
\toprule
Name      & $D$ & $m$ & Type       & Pareto Front Shape &  Reference Point \\ \midrule
RE21 (Four bar truss design)  & 4            & 2          & Continuous &     Convex &   (3144.44, 0.05)    \\
RE22 (Reinforced concrete beam design)  & 3            & 2          & Mixed &     Mixed &   (829.08, 2407217.25)    \\
RE23 (Pressure vessel design) & 4            & 2          & Mixed      &      Mixed, Disconnected   &(713710.88, 1288669.78)     \\
RE24 (Hatch cover design) & 2            & 2          & Continuous      &      Convex   &(5997.83,    43.67)     \\
RE25 (Coil compression spring design) & 3            & 2          & Mixed      &      Mixed, Disconnected   &(124.79, 10038735.00)     \\
RE31 (Two bar truss design) & 3            & 3          & Continuous      &     Unknown   &(808.85, 6893375.82, 6793450.00)     \\
RE32 (Welded beam design) & 4            & 3          & Continuous      &     Unknown   &(290.66, 16552.46, 388265024.00)     \\
RE33 (Disc brake design)                   &          4    &     3       &      Continuous      &   Unknown   & (8.01,    8.84, 2343.30)       \\
RE34 (Vehicle crashworthiness design)                    &         5     &       3     &        Continuous    &   Unknown &(1702.52, 11.68, 0.26)          \\
RE35 (Speed reducer design)                    &         7     &       3     &        Mixed    &   Unknown &(7050.79, 1696.67,  397.83)          \\
RE36 (Gear train design)                    &         4     &       3     &        Integer    &   Concave, Disconnected &(10.21, 60.00         , 0.97)          \\
RE37 (Rocket injector design)                    &     4         &     3      &      Continuous      &   Unknown   &(0.99, 0.96, 0.99)        \\
RE41 (Car side impact design)                     &  7            &        4    &      Continuous      &      Unknown &(42.65,  4.43, 13.08, 13.45)       \\
RE42 (Conceptual marine design)                     &  6            &        4    &      Continuous      &      Unknown &(-26.39, 19904.90, 28546.79, 14.98)       \\
RE61 (Water resource planning)                   &  3            &     6       &     Continuous       & Unknown  &(83060.03, 1350.00, 2853469.06, \newline 16027067.60, 357719.74, 99660.36)           \\ \bottomrule
\end{tabular}}\label{table:re}
\end{table}

\section{Detailed Experiments}\label{app:results}

\subsection{Additional Results}\label{app:additional results}

\textbf{Analysis on data pruning}
We first conduct ablation studies of data pruning on the Multi-Head model, as shown in Table~\ref{table:data-pruning-ab-results}.
\fix{Although data pruning can alleviate the issue of model collapse in some problems, it does not consistently lead to improvements in all cases. Due to the severe impact of model collapse (sometimes resulting in only one solution), we default to using data pruning for all advanced methods. As mentioned in the main paper, exploring methods to mitigate model collapse is an important future direction in offline MOO.}

\begin{table}[htbp]
\centering
\caption{Average rank of Multi-Head and Multiple Models w/ and w/o data pruning on each type of task in Off-MOO-Bench.}
\resizebox{\linewidth}{!}{
\begin{tabular}{c|cccccc|c}
\toprule
Methods                & Synthetic       & MO-NAS        & MORL            & MOCO            & Sci-Design     & RE              & Average Rank     \\ \midrule
$\mathcal{D}$(best) & 4.34 $\pm$ 0.03 & 4.74 $\pm$ 0.21 & 2.75 $\pm$ 1.25 & \textbf{1.21 $\pm$ 0.14} & 3.12 $\pm$ 0.12 & 4.07 $\pm$ 0.13 & 3.67 $\pm$ 0.15 \\
Multi-Head & 2.66 $\pm$ 0.16 & 2.66 $\pm$ 0.24 & \textbf{2.25 $\pm$ 0.75} & 3.71 $\pm$ 0.00 & 3.38 $\pm$ 0.00 & \underline{2.73 $\pm$ 0.00} & 2.90 $\pm$ 0.12 \\
Multi-Head + Data Pruning & 3.25 $\pm$ 0.12 & \textbf{2.16 $\pm$ 0.00} & 4.00 $\pm$ 1.00 & 3.31 $\pm$ 0.08 & \underline{2.38 $\pm$ 0.25} & 3.12 $\pm$ 0.12 & \underline{2.90 $\pm$ 0.02} \\
Multiple Models & \textbf{2.12 $\pm$ 0.06} & \underline{2.61 $\pm$ 0.08} & 3.25 $\pm$ 0.25 & \underline{3.07 $\pm$ 0.29} & \textbf{2.00 $\pm$ 0.25} & \textbf{1.90 $\pm$ 0.03} & \textbf{2.41 $\pm$ 0.09} \\
Multiple Models + Data Pruning & \underline{2.62 $\pm$ 0.06} & 2.84 $\pm$ 0.11 & \underline{2.75 $\pm$ 0.25} & 3.54 $\pm$ 0.23 & 3.83 $\pm$ 0.17 & 2.85 $\pm$ 0.00 & 2.94 $\pm$ 0.10 \\
\bottomrule
\end{tabular}}\label{table:data-pruning-ab-results}
\end{table}

\begin{figure}[htbp]\centering
\includegraphics[width=0.7\linewidth]{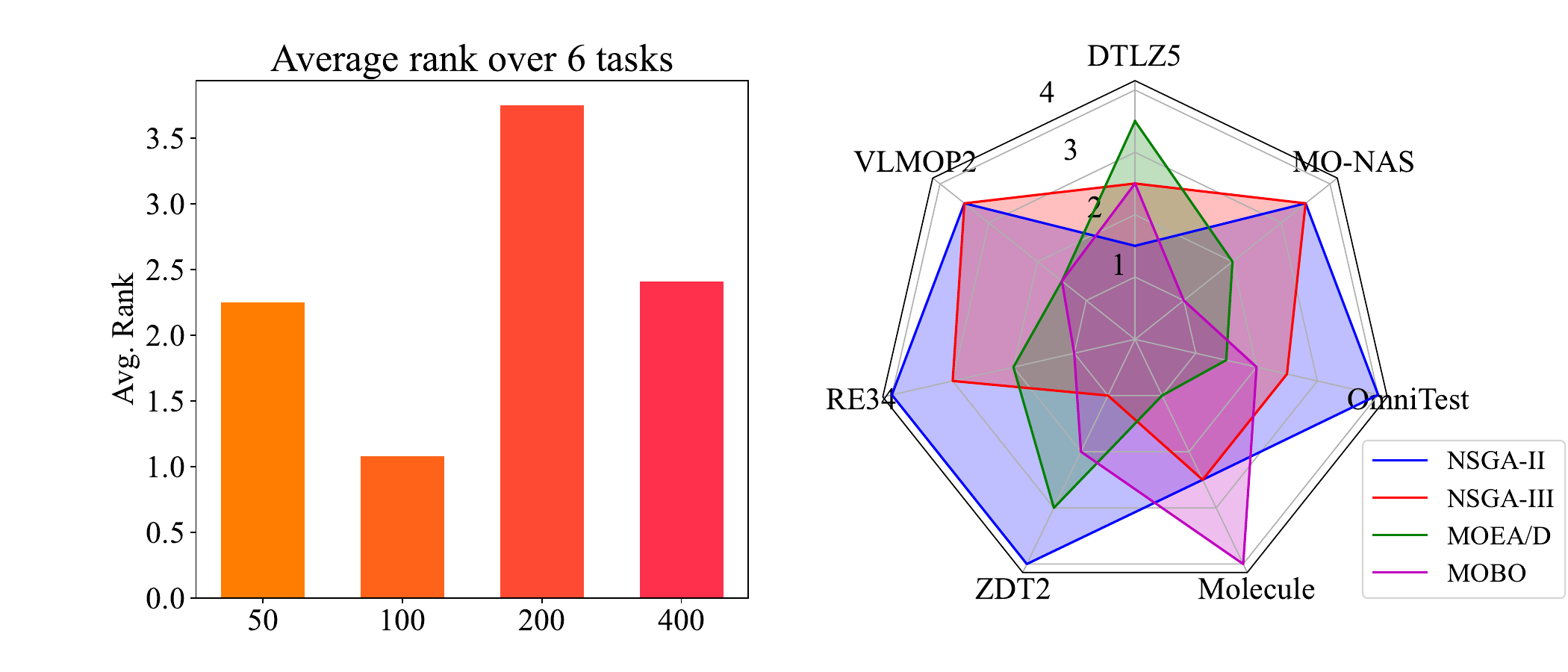}\vspace{-1em}
\caption{(a) The average rank of MOBO with different number of initial data points on six tasks. (b) The performance of four search algorithms on seven tasks.}\label{fig:mobo-ablate}
\end{figure}

\textbf{Analysis of the volume of data needed for MOBO.} As we discussed before, the number of data points for GP is important due to the complexity of learning a GP. Here, we test the influence of the different number of data points, i.e., 50, 100, 200, and 400, on six randomly selected tasks. As shown in Figure~\ref{fig:mobo-ablate} (a), 100 is a proper value. Thus, we use 100 for MOBO in our experiments on all the tasks.

\textbf{Influence of the search algorithms.} We compare four search algorithms on seven tasks, i.e., NSGA-II, MOEA/D, NSGA-III, and MOBO, as shown in Figure~\ref{fig:mobo-ablate} (b). Their average ranks are 3.28, 2.07, 2.64, and 2.00, respectively. 
Although NSGA-II has the worst average ranking, we choose to use it as the default search algorithm due to its ease of use and popularity. 
These results also show that if a search algorithm specifically designed for offline MOO is implemented, the performance can be further improved, which is an interesting future work.

\subsection{Detailed Results}
\fix{Here, we provide the detailed results on different tasks. We provide results for each type of task with 256 solutions and 100th percentile evaluations. Additionally, we provide results for each type of task with 256 solutions and 50th percentile evaluations to demonstrate the robustness of the algorithms, and results with 32 solutions and 50th percentile evaluations to show the performance under the low-budget settings. 
Considering the three settings, Multiple Models + IOM, Multiple Models, and Multi-head Model are the top three performing algorithms, with average rankings of 4.91, 5.25, and 7.16, respectively. 
% Multi-Head + GradNorm, Multiple Models + ICT, and three MOBO methods are worse than $\mathcal{D}$(best). 
Note that $\mathcal{D}$(best) achieves the best average rank on MORL, MOCO, and Sci-Design tasks on the 256 solutions with 50th percentile evaluations settings, underscoring the need for further enhancements in the robustness of offline MOO methods in these challenging tasks.

The average rank is calculated as follows: For each type of task (e.g., synthetic functions), we first determine the rankings for all methods across all sub-tasks (e.g., DTLZ1 and ZDT1) within it. After computing the six rankings for all methods, we average these values to report the average ranking of each method.}

\begin{table}[htbp]
\centering
\caption{Hypervolume results for synthetic functions with 256 solutions and 100th percentile evaluations. For each task, algorithms within one standard deviation of having the highest performance are \textbf{bolded}.}
\resizebox{\linewidth}{!}{
\begin{tabular}{c|ccccccccccccccccccc}
\toprule
Methods & DTLZ1 & DTLZ2 & DTLZ3 & DTLZ4 & DTLZ5 & DTLZ6 & DTLZ7 & OmniTest & VLMOP1 & VLMOP2 & VLMOP3 & ZDT1 & ZDT2 & ZDT3 & ZDT4 & ZDT6 \\

\midrule
$\mathcal{D}$(best) & 10.43 & 9.43 & 9.71 & \textbf{10.76} & 9.06 & 8.20 & 8.32 & 3.87 & 0.08 & 1.64 & 45.14 & 4.04 & 4.70 & 5.05 & \textbf{5.46} & 4.76 \\
End-to-End & 10.12 $\pm$ 0.02 & \textbf{10.65 $\pm$ 0.00} & 10.65 $\pm$ 0.00 & \textbf{10.70 $\pm$ 0.05} & \textbf{10.65 $\pm$ 0.00} & 10.65 $\pm$ 0.00 & 10.70 $\pm$ 0.01 & 4.35 $\pm$ 0.00 & 2.57 $\pm$ 2.26 & 4.24 $\pm$ 0.01 & 46.93 $\pm$ 0.00 & 2.69 $\pm$ 0.00 & 3.21 $\pm$ 0.00 & 5.50 $\pm$ 0.04 & 3.12 $\pm$ 0.09 & \textbf{4.92 $\pm$ 0.00} \\
End-to-End + GradNorm & 10.65 $\pm$ 0.00 & \textbf{10.65 $\pm$ 0.00} & 10.65 $\pm$ 0.00 & \textbf{10.76 $\pm$ 0.00} & 10.54 $\pm$ 0.09 & 10.64 $\pm$ 0.00 & 10.71 $\pm$ 0.00 & 3.76 $\pm$ 0.03 & 2.33 $\pm$ 2.33 & 2.79 $\pm$ 1.34 & 42.23 $\pm$ 0.98 & 4.77 $\pm$ 0.01 & 5.63 $\pm$ 0.02 & 5.27 $\pm$ 0.03 & 3.23 $\pm$ 0.03 & 3.81 $\pm$ 1.02 \\
End-to-End + PcGrad & 10.65 $\pm$ 0.00 & \textbf{10.65 $\pm$ 0.00} & 10.65 $\pm$ 0.00 & \textbf{10.70 $\pm$ 0.05} & 9.02 $\pm$ 0.10 & 9.45 $\pm$ 0.15 & 10.52 $\pm$ 0.00 & \textbf{4.35 $\pm$ 0.00} & 2.57 $\pm$ 2.26 & 4.14 $\pm$ 0.07 & 46.79 $\pm$ 0.06 & 4.84 $\pm$ 0.01 & 5.70 $\pm$ 0.01 & 5.45 $\pm$ 0.00 & 3.12 $\pm$ 0.01 & 2.04 $\pm$ 0.22 \\
Multi-Head & 10.38 $\pm$ 0.25 & \textbf{10.65 $\pm$ 0.00} & \textbf{10.65 $\pm$ 0.00} & \textbf{10.70 $\pm$ 0.05} & \textbf{10.65 $\pm$ 0.00} & 10.65 $\pm$ 0.00 & \textbf{10.63 $\pm$ 0.11} & 4.30 $\pm$ 0.05 & 2.57 $\pm$ 2.26 & 4.26 $\pm$ 0.00 & 46.92 $\pm$ 0.02 & 2.69 $\pm$ 0.00 & \textbf{4.48 $\pm$ 1.27} & 5.50 $\pm$ 0.04 & 3.23 $\pm$ 0.16 & 4.91 $\pm$ 0.00 \\
\bottomrule
\end{tabular}}\label{tab:syn-256-100}
\end{table}

\begin{table}[htbp]
\centering
\caption{Hypervolume results for MO-NAS with 256 solutions and 100th percentile evaluations. For each task, algorithms within one standard deviation of having the highest performance are \textbf{bolded}}
% \usepackage{booktabs}
\resizebox{\linewidth}{!}{
\begin{tabular}{c|ccccccccccccccccccc}
\toprule
Methods & C-10/MOP1 & C-10/MOP2 & C-10/MOP3 & C-10/MOP4 & C-10/MOP5 & C-10/MOP6 & C-10/MOP7 & C-10/MOP8 & C-10/MOP9 & IN-1K/MOP1 & IN-1K/MOP2 & IN-1K/MOP3 & IN-1K/MOP4 & IN-1K/MOP5 & IN-1K/MOP6 & IN-1K/MOP7 & IN-1K/MOP8 & IN-1K/MOP9 & NasBench201-Test \\

\midrule
$\mathcal{D}$(best) & 4.78 & 10.48 & 9.72 & 21.15 & 40.51 & 92.43 & 358.27 & 4.55 & 10.59 & 4.97 & 5.00 & 11.21 & 16.63 & 17.26 & 44.43 & 4.69 & 11.42 & 21.50 & 10.07 \\
End-to-End & 4.84 $\pm$ 0.00 & 10.49 $\pm$ 0.02 & 10.84 $\pm$ 0.01 & 26.25 $\pm$ 0.07 & 50.20 $\pm$ 0.03 & 112.47 $\pm$ 0.01 & 523.26 $\pm$ 0.11 & 4.56 $\pm$ 0.07 & 10.50 $\pm$ 0.15 & 4.85 $\pm$ 0.07 & 4.70 $\pm$ 0.02 & 10.88 $\pm$ 0.10 & 17.07 $\pm$ 0.03 & \textbf{17.93 $\pm$ 0.13} & \textbf{45.89 $\pm$ 0.16} & 5.17 $\pm$ 0.02 & 11.15 $\pm$ 0.18 & 23.35 $\pm$ 0.30 & 10.30 $\pm$ 0.03 \\
End-to-End + GradNorm & 4.84 $\pm$ 0.00 & 10.54 $\pm$ 0.01 & \textbf{10.86 $\pm$ 0.01} & 26.46 $\pm$ 0.04 & \textbf{50.64 $\pm$ 0.00} & 112.27 $\pm$ 0.03 & 525.63 $\pm$ 0.10 & 4.56 $\pm$ 0.08 & 10.48 $\pm$ 0.23 & 5.21 $\pm$ 0.00 & 5.04 $\pm$ 0.02 & 10.87 $\pm$ 0.05 & 16.93 $\pm$ 0.07 & 17.80 $\pm$ 0.09 & 45.40 $\pm$ 0.12 & 4.96 $\pm$ 0.05 & 10.73 $\pm$ 0.08 & 22.55 $\pm$ 0.13 & 9.94 $\pm$ 0.01 \\
End-to-End + PcGrad & 4.84 $\pm$ 0.00 & 10.50 $\pm$ 0.01 & 10.81 $\pm$ 0.00 & 26.30 $\pm$ 0.04 & 50.54 $\pm$ 0.08 & 112.15 $\pm$ 0.01 & 526.46 $\pm$ 0.12 & 4.58 $\pm$ 0.04 & 11.10 $\pm$ 0.04 & 5.27 $\pm$ 0.04 & 5.20 $\pm$ 0.03 & 11.11 $\pm$ 0.03 & 16.82 $\pm$ 0.01 & 17.63 $\pm$ 0.01 & 44.47 $\pm$ 0.04 & 5.03 $\pm$ 0.07 & 11.20 $\pm$ 0.20 & 23.43 $\pm$ 0.24 & 10.45 $\pm$ 0.09 \\
\bottomrule
\end{tabular}}\label{tab:nas-256-100}
\end{table}

\begin{table}[htbp]
\centering
\caption{Hypervolume results for MORL with 256 solutions and 100th percentile evaluations. For each task, algorithms within one standard deviation of having the highest performance are \textbf{bolded}.}
% \usepackage{booktabs}

\begin{tabular}{c|ccccccccccccccccccc}
\toprule
Methods  & MO-Hopper & MO-Swimmer \\

\midrule
$\mathcal{D}$(best) & 4.21 & 2.85 \\
End-to-End & 4.76 $\pm$ 0.25 & 2.77 $\pm$ 0.03 \\
End-to-End + GradNorm & \textbf{5.02 $\pm$ 0.04} & 2.90 $\pm$ 0.07 \\
End-to-End + PcGrad & 4.60 $\pm$ 0.27 & 2.49 $\pm$ 0.05 \\
Multi-Head & 4.57 $\pm$ 0.28 & 2.91 $\pm$ 0.04 \\
Multi-Head + GradNorm & 3.78 $\pm$ 0.05 & 2.69 $\pm$ 0.24 \\
Multi-Head + PcGrad & 4.27 $\pm$ 0.61 & 2.49 $\pm$ 0.25 \\
Multiple Models & 4.58 $\pm$ 0.19 & 2.60 $\pm$ 0.15 \\
Multiple Models + COMs & 4.84 $\pm$ 0.17 & 2.71 $\pm$ 0.04 \\
Multiple Models + RoMA & \textbf{5.23 $\pm$ 0.23} & 2.78 $\pm$ 0.20 \\
Multiple Models + IOM & \textbf{5.32 $\pm$ 0.49} & 2.94 $\pm$ 0.11 \\
Multiple Models + ICT & 4.67 $\pm$ 0.12 & \textbf{3.11 $\pm$ 0.08} \\
Multiple Models + Tri-Mentoring & 4.93 $\pm$ 0.11 & 2.82 $\pm$ 0.10 \\
MOBO & 4.43 $\pm$ 0.08 & 2.61 $\pm$ 0.02 \\
MOBO-$q$ParEGO & N/A & N/A \\
MOBO-JES & N/A & N/A \\ \bottomrule
\end{tabular}\label{tab:morl-256-100}
\end{table}

\begin{table}[htbp]
\centering
\caption{Hypervolume results for MOCO with 256 solutions and 100th percentile evaluations. For each task, algorithms within one standard deviation of having the highest performance are \textbf{bolded}.}
% \usepackage{booktabs}
\resizebox{\linewidth}{!}{
\begin{tabular}{c|ccccccccccccccccccc}
\toprule
Methods & Bi-CVRP-20 & Bi-CVRP-50 & Bi-CVRP-100 & Bi-KP-50 & Bi-KP-100 & Bi-KP-200 & Bi-TSP-20 & Bi-TSP-50 & Bi-TSP-100 & Bi-TSP-500 & Tri-TSP-20 & Tri-TSP-50 & Tri-TSP-100 & MO-Portfolio \\
\midrule
$\mathcal{D}$(best) & \textbf{5.37} & \textbf{5.11} & \textbf{4.93} & 3.00 & \textbf{3.45} & \textbf{4.68} & \textbf{5.05} & \textbf{4.89} & 4.55 & \textbf{4.52} & \textbf{11.88} & 9.82 & 9.36 & \textbf{3.78} \\
End-to-End & 4.85 $\pm$ 0.18 & 4.70 $\pm$ 0.17 & 4.91 $\pm$ 0.02 & 2.98 $\pm$ 0.00 & 3.02 $\pm$ 0.00 & 3.71 $\pm$ 0.47 & 3.64 $\pm$ 0.08 & 4.57 $\pm$ 0.08 & \textbf{4.60 $\pm$ 0.01} & 4.20 $\pm$ 0.29 & 6.96 $\pm$ 0.52 & 4.67 $\pm$ 0.17 & 9.26 $\pm$ 0.23 & 3.07 $\pm$ 0.16 \\
End-to-End + GradNorm & 4.77 $\pm$ 0.06 & 3.68 $\pm$ 0.20 & 3.84 $\pm$ 0.05 & 3.06 $\pm$ 0.00 & 2.84 $\pm$ 0.14 & 3.25 $\pm$ 0.01 & 3.12 $\pm$ 0.13 & 4.33 $\pm$ 0.00 & 4.04 $\pm$ 0.06 & 4.45 $\pm$ 0.02 & 7.84 $\pm$ 0.17 & 8.96 $\pm$ 0.07 & 9.41 $\pm$ 0.04 & 3.28 $\pm$ 0.15 \\
End-to-End + PcGrad & 4.56 $\pm$ 0.06 & 4.32 $\pm$ 0.09 & 4.15 $\pm$ 0.34 & \textbf{3.16 $\pm$ 0.02} & 3.08 $\pm$ 0.03 & 3.52 $\pm$ 0.27 & 3.08 $\pm$ 0.03 & 4.68 $\pm$ 0.03 & 4.55 $\pm$ 0.02 & 4.17 $\pm$ 0.12 & 10.83 $\pm$ 0.03 & 8.15 $\pm$ 0.39 & 9.68 $\pm$ 0.15 & 3.08 $\pm$ 0.05 \\
Multi-Head & 4.00 $\pm$ 0.06 & 4.03 $\pm$ 0.26 & \textbf{4.78 $\pm$ 0.16} & \textbf{3.11 $\pm$ 0.06} & 2.88 $\pm$ 0.04 & 4.03 $\pm$ 0.41 & 3.36 $\pm$ 0.14 & 4.60 $\pm$ 0.09 & 4.04 $\pm$ 0.00 & 3.70 $\pm$ 0.27 & 6.31 $\pm$ 0.14 & 6.43 $\pm$ 0.42 & 8.60 $\pm$ 0.13 & 3.18 $\pm$ 0.04 \\
Multi-Head + GradNorm & 4.50 $\pm$ 0.51 & 3.69 $\pm$ 0.12 & 4.35 $\pm$ 0.57 & 2.88 $\pm$ 0.07 & 2.47 $\pm$ 0.01 & 2.98 $\pm$ 0.41 & 4.39 $\pm$ 0.17 & 3.40 $\pm$ 0.02 & 3.49 $\pm$ 0.39 & 3.83 $\pm$ 0.02 & 7.84 $\pm$ 0.20 & 6.48 $\pm$ 0.65 & 8.48 $\pm$ 0.23 & 3.11 $\pm$ 0.11 \\
\bottomrule
\end{tabular}}\label{tab:moco-256-100}
\end{table}

\begin{table}[htbp]
\centering
\caption{Hypervolume results for scientific design with 256 solutions and 100th percentile evaluations. For each task, algorithms within one standard deviation of having the highest performance are \textbf{bolded}.}
% \usepackage{booktabs}

\begin{tabular}{c|ccccccccccccccccccc}
\toprule
Methods  & Molecule & Regex & RFP & ZINC \\

\midrule
$\mathcal{D}$(best) & 2.26  & 2.82  & 3.36  & 4.01\\
End-to-End & \textbf{2.30 $\pm$ 0.48} & 2.80 $\pm$ 0.00 & 3.80 $\pm$ 0.04 & 4.17 $\pm$ 0.00 \\
End-to-End + GradNorm & 1.10 $\pm$ 0.03 & 2.80 $\pm$ 0.00 & 4.11 $\pm$ 0.30 & 4.17 $\pm$ 0.00 \\
End-to-End + PcGrad & 1.54 $\pm$ 0.53 & 2.80 $\pm$ 0.00 & 3.84 $\pm$ 0.05 & 4.16 $\pm$ 0.08 \\
Multi-Head & 2.08 $\pm$ 0.00 & 2.80 $\pm$ 0.00 & 3.75 $\pm$ 0.00 & 4.16 $\pm$ 0.00 \\
Multi-Head + GradNorm & 1.62 $\pm$ 0.61 & 2.38 $\pm$ 0.00 & 4.08 $\pm$ 0.32 & 4.21 $\pm$ 0.05 \\
Multi-Head + PcGrad & 1.22 $\pm$ 0.10 & 2.80 $\pm$ 0.00 & 4.19 $\pm$ 0.22 & 4.12 $\pm$ 0.02 \\
Multiple Models & \textbf{2.78 $\pm$ 0.00} & 2.80 $\pm$ 0.00 & \textbf{4.40 $\pm$ 0.02} & 4.16 $\pm$ 0.00 \\
Multiple Models + COMs & 2.30 $\pm$ 0.00 & 2.21 $\pm$ 0.17 & \textbf{4.14 $\pm$ 0.35} & 4.12 $\pm$ 0.05 \\
Multiple Models + RoMA & 1.65 $\pm$ 0.02 & 2.80 $\pm$ 0.00 & \textbf{4.13 $\pm$ 0.29} & 4.16 $\pm$ 0.01 \\
Multiple Models + IOM & 1.75 $\pm$ 0.33 & 2.80 $\pm$ 0.00 & \textbf{4.13 $\pm$ 0.28} & 4.17 $\pm$ 0.00 \\
Multiple Models + ICT & 1.37 $\pm$ 0.17 & 2.80 $\pm$ 0.00 & \textbf{4.41 $\pm$ 0.00} & 4.10 $\pm$ 0.07 \\
Multiple Models + Tri-Mentoring & 2.03 $\pm$ 0.00 & 2.80 $\pm$ 0.00 & \textbf{4.12 $\pm$ 0.29} & 4.06 $\pm$ 0.01 \\
MOBO & 2.22 $\pm$ 0.08 & \textbf{5.12 $\pm$ 0.17} & 3.74 $\pm$ 0.00 & \textbf{4.26 $\pm$ 0.00} \\
MOBO-$q$ParEGO & 2.12 $\pm$ 0.04 & 4.26 $\pm$ 0.25 & 3.33 $\pm$ 0.00 & 4.05 $\pm$ 0.02 \\
MOBO-JES & \textbf{2.10 $\pm$ 1.04} & N/A & N/A & N/A \\ \bottomrule
\end{tabular}\label{tab:sci-design-256-100}
\end{table}

\begin{table}[htbp]
\centering
\caption{Hypervolume results for RE with 256 solutions and 100th percentile evaluations. For each task, algorithms within one standard deviation of having the highest performance are \textbf{bolded}.}
% \usepackage{booktabs}
\resizebox{\linewidth}{!}{
\begin{tabular}{c|ccccccccccccccccccc}
\toprule

Methods & RE21 & RE22 & RE23 & RE24 & RE25 & RE31 & RE32 & RE33 & RE34 & RE35 & RE36 & RE37 & RE41 & RE42 & RE61 \\

\midrule
$\mathcal{D}$(best) & 4.23 & 4.78 & 4.75 & 4.59 & 4.79 & 10.23 & 10.53 & 10.59 & 48.06 & 10.96 & 7.57 & 4.72 & 36.17 & 12.53 & 135.87 \\
End-to-End & 4.42 $\pm$ 0.00 & 4.84 $\pm$ 0.00 & \textbf{4.84 $\pm$ 0.00} & 4.38 $\pm$ 0.00 & 4.84 $\pm$ 0.00 & 10.56 $\pm$ 0.00 & 10.64 $\pm$ 0.00 & 10.69 $\pm$ 0.00 & 52.86 $\pm$ 0.05 & 11.69 $\pm$ 0.00 & 9.25 $\pm$ 1.00 & 6.21 $\pm$ 0.00 & 44.13 $\pm$ 0.02 & 20.04 $\pm$ 0.01 & \textbf{144.30 $\pm$ 0.02} \\
End-to-End + GradNorm & 4.81 $\pm$ 0.01 & 4.84 $\pm$ 0.00 & 2.64 $\pm$ 0.00 & 4.38 $\pm$ 0.00 & \textbf{4.84 $\pm$ 0.00} & 10.65 $\pm$ 0.00 & 10.63 $\pm$ 0.00 & 9.90 $\pm$ 0.00 & 51.38 $\pm$ 0.03 & 11.52 $\pm$ 0.00 & 9.16 $\pm$ 0.02 & 6.22 $\pm$ 0.01 & 41.26 $\pm$ 0.81 & 13.46 $\pm$ 0.00 & 141.37 $\pm$ 0.03 \\
End-to-End + PcGrad & 4.90 $\pm$ 0.05 & 4.84 $\pm$ 0.00 & 4.84 $\pm$ 0.00 & 4.38 $\pm$ 0.00 & 4.60 $\pm$ 0.24 & \textbf{10.65 $\pm$ 0.00} & 10.65 $\pm$ 0.00 & 10.41 $\pm$ 0.07 & 52.83 $\pm$ 0.08 & 11.68 $\pm$ 0.02 & 10.02 $\pm$ 0.00 & 5.52 $\pm$ 0.00 & 43.53 $\pm$ 0.37 & 14.27 $\pm$ 0.12 & 142.98 $\pm$ 0.38 \\
Multi-Head & 4.57 $\pm$ 0.15 & 4.84 $\pm$ 0.00 & 4.74 $\pm$ 0.00 & 4.78 $\pm$ 0.00 & 4.60 $\pm$ 0.24 & 10.65 $\pm$ 0.00 & 10.64 $\pm$ 0.00 & 10.69 $\pm$ 0.00 & 52.93 $\pm$ 0.01 & 11.74 $\pm$ 0.00 & 6.76 $\pm$ 0.00 & 5.78 $\pm$ 0.05 & 44.06 $\pm$ 0.02 & \textbf{20.71 $\pm$ 0.29} & 141.28 $\pm$ 1.13 \\
Multi-Head + GradNorm & 4.91 $\pm$ 0.00 & 4.83 $\pm$ 0.01 & 4.49 $\pm$ 0.09 & 2.64 $\pm$ 0.00 & 3.95 $\pm$ 0.00 & 10.65 $\pm$ 0.00 & 10.63 $\pm$ 0.00 & 5.85 $\pm$ 0.00 & 52.84 $\pm$ 0.00 & 11.52 $\pm$ 0.00 & 0.02 $\pm$ 0.00 & 6.36 $\pm$ 0.01 & 43.77 $\pm$ 0.09 & 19.01 $\pm$ 0.04 & 143.82 $\pm$ 0.20 \\
\bottomrule
\end{tabular}}\label{tab:re-256-100}
\end{table}

\begin{table}[htbp]
\centering
\caption{Average rank of different offline MOO methods on each type of task with 256 solutions and 50th percentile evaluations, where the best and runner-up results are \textbf{bolded} and \underline{underlined}, respectively.}
% \usepackage{booktabs}
\resizebox{\linewidth}{!}{
\begin{tabular}{c|cccccc|c}
\toprule
Methods  & Synthetic & MO-NAS & MORL & MOCO & Sci-Design & RE & Average Rank \\

\midrule
$\mathcal{D}$(best) & 9.20 $\pm$ 0.42 & 9.58 $\pm$ 0.11 & \textbf{1.75 $\pm$ 0.25} & \textbf{1.07 $\pm$ 0.00} & \textbf{1.88 $\pm$ 0.62} & 10.50 $\pm$ 0.30 & 7.22 $\pm$ 0.14 \\
End-to-End & 8.50 $\pm$ 0.38 & 7.53 $\pm$ 0.21 & 9.50 $\pm$ 2.00 & 7.38 $\pm$ 0.38 & 9.25 $\pm$ 1.00 & 6.33 $\pm$ 0.13 & 7.55 $\pm$ 0.19 \\
End-to-End + GradNorm & 9.34 $\pm$ 0.34 & 8.34 $\pm$ 0.29 & 5.75 $\pm$ 1.75 & 8.08 $\pm$ 0.92 & 8.00 $\pm$ 0.25 & 9.67 $\pm$ 0.00 & 8.76 $\pm$ 0.12 \\
End-to-End + PcGrad & 8.61 $\pm$ 0.17 & 8.08 $\pm$ 0.34 & 10.00 $\pm$ 1.00 & 8.50 $\pm$ 0.43 & 7.62 $\pm$ 0.12 & 7.77 $\pm$ 0.57 & 8.29 $\pm$ 0.18 \\
Multi-Head & 7.47 $\pm$ 1.09 & 5.16 $\pm$ 0.05 & 5.75 $\pm$ 1.75 & 9.46 $\pm$ 0.39 & 7.25 $\pm$ 0.25 & 7.90 $\pm$ 0.50 & 7.19 $\pm$ 0.48 \\
Multi-Head + GradNorm & 8.48 $\pm$ 0.52 & 10.46 $\pm$ 0.20 & 10.75 $\pm$ 2.75 & 8.64 $\pm$ 0.43 & 12.25 $\pm$ 0.25 & 10.07 $\pm$ 0.40 & 9.67 $\pm$ 0.23 \\
Multi-Head + PcGrad & 7.95 $\pm$ 0.52 & 8.97 $\pm$ 0.66 & 7.25 $\pm$ 2.75 & 8.49 $\pm$ 0.28 & 11.79 $\pm$ 0.54 & 8.43 $\pm$ 0.10 & 8.55 $\pm$ 0.33 \\
\bottomrule
\end{tabular}}\label{tab:avg-rank-256-50}
\end{table}

\begin{table}[htbp]
\centering
\caption{Hypervolume results for synthetic functions with 256 solutions and 50th percentile evaluations. For each task, algorithms within one standard deviation of having the highest performance are \textbf{bolded}.}
% \usepackage{booktabs}
\resizebox{\linewidth}{!}{
\begin{tabular}{c|ccccccccccccccccccc}
\toprule
Methods  & DTLZ1 & DTLZ2 & DTLZ3 & DTLZ4 & DTLZ5 & DTLZ6 & DTLZ7 & OmniTest & VLMOP1 & VLMOP2 & VLMOP3 & ZDT1 & ZDT2 & ZDT3 & ZDT4 & ZDT6 \\

\midrule
$\mathcal{D}$(best) & 10.43 & 9.43 & 9.71 & \textbf{10.76} & 9.06 & 8.20 & 8.32 & 3.87 & 0.08 & 1.64 & 45.14 & 4.04 & 4.70 & 5.05 & \textbf{5.46} & 4.76 \\
End-to-End & 10.06 $\pm$ 0.00 & \textbf{10.65 $\pm$ 0.00} & \textbf{10.65 $\pm$ 0.00} & 9.98 $\pm$ 0.35 & \textbf{8.63 $\pm$ 2.01} & 9.42 $\pm$ 0.00 & 6.37 $\pm$ 0.07 & \textbf{4.35 $\pm$ 0.00} & 0.00 $\pm$ 0.00 & 4.18 $\pm$ 0.02 & 46.76 $\pm$ 0.09 & 2.69 $\pm$ 0.00 & 3.21 $\pm$ 0.00 & 5.46 $\pm$ 0.00 & 3.04 $\pm$ 0.02 & \textbf{4.87 $\pm$ 0.02} \\
End-to-End + GradNorm & \textbf{10.65 $\pm$ 0.00} & 10.36 $\pm$ 0.28 & 10.63 $\pm$ 0.02 & \textbf{10.28 $\pm$ 0.48} & \textbf{8.50 $\pm$ 2.07} & 8.69 $\pm$ 0.68 & \textbf{8.62 $\pm$ 2.08} & 2.32 $\pm$ 0.04 & 0.00 $\pm$ 0.00 & 2.67 $\pm$ 1.21 & 38.20 $\pm$ 0.17 & 4.76 $\pm$ 0.00 & 4.01 $\pm$ 0.17 & 5.27 $\pm$ 0.03 & 3.02 $\pm$ 0.02 & 2.55 $\pm$ 0.23 \\
End-to-End + PcGrad & 10.62 $\pm$ 0.02 & 8.22 $\pm$ 1.99 & 10.65 $\pm$ 0.00 & \textbf{10.76 $\pm$ 0.00} & 8.01 $\pm$ 0.07 & 7.80 $\pm$ 0.80 & 10.52 $\pm$ 0.00 & 4.32 $\pm$ 0.03 & 1.36 $\pm$ 1.36 & 4.08 $\pm$ 0.10 & 34.65 $\pm$ 0.06 & 4.37 $\pm$ 0.29 & 5.70 $\pm$ 0.01 & 4.45 $\pm$ 0.94 & 2.99 $\pm$ 0.02 & 1.87 $\pm$ 0.10 \\
Multi-Head & 10.37 $\pm$ 0.24 & 10.64 $\pm$ 0.01 & 10.63 $\pm$ 0.01 & 10.14 $\pm$ 0.19 & 6.62 $\pm$ 0.00 & 9.39 $\pm$ 0.03 & \textbf{10.61 $\pm$ 0.09} & 4.29 $\pm$ 0.05 & 0.95 $\pm$ 0.95 & 4.18 $\pm$ 0.01 & \textbf{46.78 $\pm$ 0.16} & 2.69 $\pm$ 0.00 & \textbf{4.48 $\pm$ 1.27} & 5.50 $\pm$ 0.03 & 2.94 $\pm$ 0.08 & \textbf{4.90 $\pm$ 0.00} \\
\bottomrule
\end{tabular}}\label{tab:syn-256-50}
\end{table}

\begin{table}[htbp]
\centering
\caption{Hypervolume results for MO-NAS with 256 solutions and 50th percentile evaluations. For each task, algorithms within one standard deviation of having the highest performance are \textbf{bolded}.}
% \usepackage{booktabs}
\resizebox{\linewidth}{!}{
\begin{tabular}{c|ccccccccccccccccccc}
\toprule
Methods  & C-10/MOP1 & C-10/MOP2 & C-10/MOP3 & C-10/MOP4 & C-10/MOP5 & C-10/MOP6 & C-10/MOP7 & C-10/MOP8 & C-10/MOP9 & IN-1K/MOP1 & IN-1K/MOP2 & IN-1K/MOP3 & IN-1K/MOP4 & IN-1K/MOP5 & IN-1K/MOP6 & IN-1K/MOP7 & IN-1K/MOP8 & IN-1K/MOP9 & NasBench201-Test \\

\midrule
$\mathcal{D}$(best) & 4.78 & 10.48 & 9.72 & 21.15 & 40.51 & 92.43 & 358.27 & 4.55 & 10.59 & 4.97 & 5.00 & 11.21 & 16.63 & 17.26 & 44.43 & 4.69 & 11.42 & 21.50 & 10.07 \\
End-to-End & \textbf{4.80 $\pm$ 0.00} & 10.38 $\pm$ 0.10 & 10.66 $\pm$ 0.02 & 24.71 $\pm$ 0.03 & 48.81 $\pm$ 0.31 & 109.63 $\pm$ 0.09 & 513.62 $\pm$ 1.06 & 4.41 $\pm$ 0.07 & 9.29 $\pm$ 0.16 & 4.75 $\pm$ 0.01 & 4.04 $\pm$ 0.00 & 10.27 $\pm$ 0.13 & 16.81 $\pm$ 0.01 & \textbf{17.68 $\pm$ 0.06} & \textbf{45.13 $\pm$ 0.21} & 4.66 $\pm$ 0.22 & 10.33 $\pm$ 0.25 & 22.45 $\pm$ 0.31 & 10.22 $\pm$ 0.00 \\
End-to-End + GradNorm & 4.78 $\pm$ 0.00 & 10.51 $\pm$ 0.00 & \textbf{10.72 $\pm$ 0.00} & 24.11 $\pm$ 0.34 & 49.40 $\pm$ 0.08 & 110.13 $\pm$ 0.06 & 517.76 $\pm$ 1.15 & 4.25 $\pm$ 0.23 & 9.52 $\pm$ 0.34 & 5.11 $\pm$ 0.05 & 4.96 $\pm$ 0.03 & 10.29 $\pm$ 0.03 & 15.29 $\pm$ 0.00 & 17.40 $\pm$ 0.05 & 44.18 $\pm$ 0.00 & 4.60 $\pm$ 0.15 & 10.56 $\pm$ 0.24 & 22.07 $\pm$ 0.17 & 9.17 $\pm$ 0.03 \\
End-to-End + PcGrad & \textbf{4.80 $\pm$ 0.00} & 10.49 $\pm$ 0.01 & 10.65 $\pm$ 0.01 & 23.78 $\pm$ 0.15 & 50.02 $\pm$ 0.08 & 107.50 $\pm$ 0.05 & 512.26 $\pm$ 0.83 & 3.97 $\pm$ 0.01 & 10.72 $\pm$ 0.08 & 4.72 $\pm$ 0.01 & \textbf{5.16 $\pm$ 0.04} & 10.06 $\pm$ 0.57 & 15.42 $\pm$ 0.04 & 17.35 $\pm$ 0.02 & 43.51 $\pm$ 0.02 & 4.84 $\pm$ 0.01 & 10.15 $\pm$ 0.78 & 22.68 $\pm$ 0.10 & 10.01 $\pm$ 0.22 \\
\bottomrule
\end{tabular}}\label{tab:nas-256-50}
\end{table}

\begin{table}[htbp]
\centering
\caption{Hypervolume results for MORL with 256 solutions and 50th percentile evaluations. For each task, algorithms within one standard deviation of having the highest performance are \textbf{bolded}.}
% \usepackage{booktabs}

\begin{tabular}{c|ccccccccccccccccccc}
\toprule
Methods & MO-Hopper & MO-Swimmer \\

\midrule
$\mathcal{D}$(best) & 4.21 & \textbf{2.85} \\
End-to-End & 3.68 $\pm$ 0.00 & 2.04 $\pm$ 0.10 \\
End-to-End + GradNorm & 3.94 $\pm$ 0.23 & 2.08 $\pm$ 0.02 \\
End-to-End + PcGrad & 3.72 $\pm$ 0.01 & 1.90 $\pm$ 0.05 \\
Multi-Head & 3.74 $\pm$ 0.07 & 2.66 $\pm$ 0.04 \\
Multi-Head + GradNorm & 3.67 $\pm$ 0.00 & 1.98 $\pm$ 0.12 \\
Multi-Head + PcGrad & 3.86 $\pm$ 0.18 & 2.08 $\pm$ 0.02 \\
Multiple Models & 3.76 $\pm$ 0.01 & 1.91 $\pm$ 0.02 \\
Multiple Models + COMs & 3.72 $\pm$ 0.02 & 1.98 $\pm$ 0.01 \\
Multiple Models + RoMA & \textbf{4.74 $\pm$ 0.00} & 1.95 $\pm$ 0.06 \\
Multiple Models + IOM & 4.17 $\pm$ 0.18 & 1.96 $\pm$ 0.06 \\
Multiple Models + ICT & 3.70 $\pm$ 0.01 & 2.38 $\pm$ 0.11 \\
Multiple Models + Tri-Mentoring & 3.82 $\pm$ 0.03 & 1.98 $\pm$ 0.01 \\
MOBO & 3.68 $\pm$ 0.00 & 1.49 $\pm$ 0.02 \\
MOBO-$q$ParEGO & N/A & N/A \\
MOBO-JES & N/A & N/A \\ \bottomrule
\end{tabular}\label{tab:morl-256-50}
\end{table}

\begin{table}[htbp]
\centering
\caption{Hypervolume results for MOCO with 256 solutions and 50th percentile evaluations. For each task, algorithms within one standard deviation of having the highest performance are \textbf{bolded}.}
% \usepackage{booktabs}
\resizebox{\linewidth}{!}{
\begin{tabular}{c|ccccccccccccccccccc}
\toprule
Methods & Bi-CVRP-20 & Bi-CVRP-50 & Bi-CVRP-100 & Bi-KP-50 & Bi-KP-100 & Bi-KP-200 & Bi-TSP-20 & Bi-TSP-50 & Bi-TSP-100 & Bi-TSP-500 & Tri-TSP-20 & Tri-TSP-50 & Tri-TSP-100 & MO-Portfolio\\

\midrule
$\mathcal{D}$(best) & \textbf{5.37} & \textbf{5.11} & \textbf{4.93} & \textbf{3.00} & \textbf{3.45} & \textbf{4.68} & \textbf{5.05} & \textbf{4.89} & \textbf{4.55} & \textbf{4.52} & \textbf{11.88} & \textbf{9.82} & 9.36 & \textbf{3.78} \\
End-to-End & 3.48 $\pm$ 0.08 & 3.30 $\pm$ 0.22 & 4.14 $\pm$ 0.02 & 2.40 $\pm$ 0.03 & 1.87 $\pm$ 0.00 & 3.69 $\pm$ 0.00 & 2.54 $\pm$ 0.22 & 4.12 $\pm$ 0.01 & 3.83 $\pm$ 0.58 & 3.35 $\pm$ 0.01 & 4.43 $\pm$ 0.35 & 3.69 $\pm$ 0.04 & 6.27 $\pm$ 0.56 & 2.97 $\pm$ 0.14 \\
End-to-End + GradNorm & 3.26 $\pm$ 0.05 & 2.93 $\pm$ 0.15 & 2.95 $\pm$ 0.02 & 2.23 $\pm$ 0.30 & 2.35 $\pm$ 0.22 & 1.90 $\pm$ 0.00 & 2.26 $\pm$ 0.10 & 3.21 $\pm$ 0.27 & 3.03 $\pm$ 0.18 & 4.19 $\pm$ 0.06 & 4.84 $\pm$ 0.36 & 4.66 $\pm$ 0.30 & 8.02 $\pm$ 0.11 & 3.14 $\pm$ 0.14 \\
End-to-End + PcGrad & 3.15 $\pm$ 0.07 & 3.39 $\pm$ 0.05 & 3.14 $\pm$ 0.24 & 2.02 $\pm$ 0.05 & 1.95 $\pm$ 0.04 & 2.35 $\pm$ 0.00 & 2.29 $\pm$ 0.03 & 3.86 $\pm$ 0.24 & 3.32 $\pm$ 0.02 & 3.65 $\pm$ 0.09 & 7.04 $\pm$ 1.00 & 5.06 $\pm$ 0.93 & 8.50 $\pm$ 0.48 & 1.99 $\pm$ 0.27 \\
Multi-Head & 3.09 $\pm$ 0.11 & 3.21 $\pm$ 0.01 & 3.84 $\pm$ 0.26 & 2.45 $\pm$ 0.01 & 1.75 $\pm$ 0.04 & 2.49 $\pm$ 0.00 & 2.45 $\pm$ 0.07 & 4.00 $\pm$ 0.08 & 3.25 $\pm$ 0.05 & 3.04 $\pm$ 0.09 & 3.90 $\pm$ 0.14 & 4.22 $\pm$ 0.14 & 6.29 $\pm$ 0.94 & 2.02 $\pm$ 0.22 \\
Multi-Head + GradNorm & 3.10 $\pm$ 0.14 & 3.12 $\pm$ 0.22 & 2.64 $\pm$ 0.16 & 2.21 $\pm$ 0.02 & 2.18 $\pm$ 0.17 & 2.20 $\pm$ 0.17 & 3.23 $\pm$ 0.05 & 2.80 $\pm$ 0.04 & 2.92 $\pm$ 0.11 & 3.18 $\pm$ 0.03 & 4.50 $\pm$ 0.09 & 5.43 $\pm$ 0.47 & 6.52 $\pm$ 0.51 & 3.06 $\pm$ 0.09 \\
\bottomrule
\end{tabular}}\label{tab:moco-256-50}
\end{table}

\begin{table}[htbp]
\centering
\caption{Hypervolume results for scientific design with 256 solutions and 50th percentile evaluations. For each task, algorithms within one standard deviation of having the highest performance are \textbf{bolded}.}
% \usepackage{booktabs}

\begin{tabular}{c|ccccccccccccccccccc}
\toprule
Methods  & Molecule & Regex & RFP & ZINC \\

\midrule
$\mathcal{D}$(best) & \textbf{2.26} & 3.05 & 3.75 & \textbf{4.06} \\
End-to-End & 1.07 $\pm$ 0.07 & 2.05 $\pm$ 0.00 & 3.64 $\pm$ 0.05 & 3.95 $\pm$ 0.04 \\
End-to-End + GradNorm & 1.07 $\pm$ 0.07 & 2.05 $\pm$ 0.00 & 3.73 $\pm$ 0.04 & 3.92 $\pm$ 0.00 \\
End-to-End + PcGrad & 2.12 $\pm$ 0.04 & 2.05 $\pm$ 0.00 & 3.70 $\pm$ 0.05 & 3.89 $\pm$ 0.06 \\
Multi-Head & 2.08 $\pm$ 0.00 & 2.05 $\pm$ 0.00 & 3.74 $\pm$ 0.00 & 3.86 $\pm$ 0.02 \\
Multi-Head + GradNorm & 1.00 $\pm$ 0.00 & 2.05 $\pm$ 0.00 & 3.69 $\pm$ 0.01 & 3.82 $\pm$ 0.01 \\
Multi-Head + PcGrad & 1.00 $\pm$ 0.00 & 2.05 $\pm$ 0.00 & 3.68 $\pm$ 0.02 & 3.86 $\pm$ 0.01 \\
Multiple Models & 1.10 $\pm$ 0.09 & 2.05 $\pm$ 0.00 & 3.70 $\pm$ 0.01 & 3.84 $\pm$ 0.00 \\
Multiple Models + COMs & 1.76 $\pm$ 0.14 & 2.38 $\pm$ 0.33 & 3.70 $\pm$ 0.00 & 3.86 $\pm$ 0.02 \\
Multiple Models + RoMA & 1.03 $\pm$ 0.00 & 2.05 $\pm$ 0.00 & \textbf{3.79 $\pm$ 0.04} & 3.91 $\pm$ 0.02 \\
Multiple Models + IOM & 1.02 $\pm$ 0.01 & 2.05 $\pm$ 0.00 & \textbf{3.76 $\pm$ 0.03} & 3.91 $\pm$ 0.02 \\
Multiple Models + ICT & 1.02 $\pm$ 0.02 & 2.05 $\pm$ 0.00 & 3.67 $\pm$ 0.00 & 3.96 $\pm$ 0.07 \\
Multiple Models + Tri-Mentoring & 1.41 $\pm$ 0.17 & 2.05 $\pm$ 0.00 & 3.75 $\pm$ 0.03 & 3.75 $\pm$ 0.00 \\
MOBO & 1.02 $\pm$ 0.02 & \textbf{3.42 $\pm$ 0.25} & 3.70 $\pm$ 0.01 & 3.90 $\pm$ 0.01 \\
MOBO-$q$ParEGO & 1.96 $\pm$ 0.12 & \textbf{3.17 $\pm$ 0.11} & 3.33 $\pm$ 0.00 & 4.00 $\pm$ 0.03 \\
MOBO-JES & 1.00 $\pm$ 0.00 & N/A & N/A & N/A \\\bottomrule
\end{tabular}\label{tab:sci-design-256-50}
\end{table}

\begin{table}[htbp]
\centering
\caption{Hypervolume results for RE with 256 solutions and 50th percentile evaluations. For each task, algorithms within one standard deviation of having the highest performance are \textbf{bolded}.}
\resizebox{\linewidth}{!}{
\begin{tabular}{c|ccccccccccccccccccc}
\toprule

Methods & RE21 & RE22 & RE23 & RE24 & RE25 & RE31 & RE32 & RE33 & RE34 & RE35 & RE36 & RE37 & RE41 & RE42 & RE61 \\

\midrule
$\mathcal{D}$(best) & 4.23 & 4.78 & 4.75 & 4.59 & 4.79 & 10.23 & 10.53 & 10.59 & 48.06 & 10.96 & 7.57 & 4.72 & 36.17 & 12.53 & 135.87 \\
End-to-End & 4.42 $\pm$ 0.00 & 4.84 $\pm$ 0.00 & 4.84 $\pm$ 0.00 & 4.38 $\pm$ 0.00 & 4.73 $\pm$ 0.04 & 10.56 $\pm$ 0.00 & 10.64 $\pm$ 0.00 & \textbf{10.68 $\pm$ 0.00} & 50.27 $\pm$ 0.40 & 11.69 $\pm$ 0.00 & 8.86 $\pm$ 0.79 & 4.67 $\pm$ 0.35 & 43.75 $\pm$ 0.07 & 19.89 $\pm$ 0.05 & \textbf{143.86 $\pm$ 0.14} \\
End-to-End + GradNorm & 4.39 $\pm$ 0.02 & 4.84 $\pm$ 0.00 & 2.64 $\pm$ 0.00 & 4.29 $\pm$ 0.00 & \textbf{4.84 $\pm$ 0.00} & 10.65 $\pm$ 0.00 & 10.61 $\pm$ 0.00 & 9.72 $\pm$ 0.03 & 47.25 $\pm$ 3.23 & 11.47 $\pm$ 0.01 & 5.74 $\pm$ 0.00 & 6.02 $\pm$ 0.07 & 40.63 $\pm$ 1.01 & 13.46 $\pm$ 0.00 & 137.59 $\pm$ 3.59 \\
End-to-End + PcGrad & \textbf{4.61 $\pm$ 0.32} & 4.52 $\pm$ 0.32 & \textbf{4.84 $\pm$ 0.00} & 4.22 $\pm$ 0.02 & 4.35 $\pm$ 0.00 & 10.65 $\pm$ 0.00 & 10.64 $\pm$ 0.00 & 9.86 $\pm$ 0.36 & 52.17 $\pm$ 0.55 & 11.67 $\pm$ 0.01 & 9.46 $\pm$ 0.03 & 4.00 $\pm$ 0.18 & 41.87 $\pm$ 0.01 & 15.81 $\pm$ 1.68 & 140.21 $\pm$ 2.41 \\
Multi-Head & 4.56 $\pm$ 0.14 & 4.83 $\pm$ 0.01 & 4.59 $\pm$ 0.10 & 4.11 $\pm$ 0.01 & 3.82 $\pm$ 0.30 & 10.64 $\pm$ 0.00 & 10.64 $\pm$ 0.00 & \textbf{10.47 $\pm$ 0.22} & 51.24 $\pm$ 0.12 & 11.73 $\pm$ 0.00 & 6.76 $\pm$ 0.00 & 4.43 $\pm$ 0.01 & 43.62 $\pm$ 0.02 & 20.20 $\pm$ 0.15 & 140.38 $\pm$ 0.56 \\
Multi-Head + GradNorm & 4.84 $\pm$ 0.01 & 3.75 $\pm$ 0.06 & 3.70 $\pm$ 0.09 & 2.64 $\pm$ 0.00 & 3.14 $\pm$ 0.01 & 10.65 $\pm$ 0.00 & 10.62 $\pm$ 0.01 & 6.12 $\pm$ 0.49 & 52.50 $\pm$ 0.14 & 11.52 $\pm$ 0.00 & 0.02 $\pm$ 0.00 & 5.90 $\pm$ 0.44 & 43.41 $\pm$ 0.16 & 15.25 $\pm$ 1.79 & 142.41 $\pm$ 0.63 \\
\bottomrule
\end{tabular}}\label{tab:re-256-50}
\end{table}

\begin{table}[htbp]
\centering
\caption{Average rank of different offline MOO methods on each type of task with 32 solutions and 100th percentile evaluations, where the best and runner-up results are \textbf{bolded} and \underline{underlined}, respectively.}
% \usepackage{booktabs}
\resizebox{\linewidth}{!}{
\begin{tabular}{c|cccccc|c}
\toprule
Methods  & Synthetic & MO-NAS & MORL & MOCO & Sci-Design & RE & Average Rank \\

\midrule
$\mathcal{D}$(best) & 12.03 $\pm$ 0.16 & 10.58 $\pm$ 0.32 & \underline{4.00 $\pm$ 0.00} & \textbf{1.07 $\pm$ 0.00} & \underline{4.25 $\pm$ 0.25} & 13.20 $\pm$ 0.07 & 8.95 $\pm$ 0.12 \\
End-to-End & 7.03 $\pm$ 0.09 & 7.35 $\pm$ 0.13 & 7.00 $\pm$ 1.00 & 5.68 $\pm$ 0.68 & \textbf{3.81 $\pm$ 0.31} & 8.47 $\pm$ 0.13 & 6.95 $\pm$ 0.13 \\
End-to-End + GradNorm & 7.80 $\pm$ 0.02 & 8.29 $\pm$ 0.24 & 6.00 $\pm$ 2.00 & 7.82 $\pm$ 0.18 & 12.25 $\pm$ 0.25 & 10.43 $\pm$ 0.50 & 8.71 $\pm$ 0.00 \\
End-to-End + PcGrad & 8.05 $\pm$ 0.67 & 7.69 $\pm$ 0.52 & 4.00 $\pm$ 0.00 & 6.99 $\pm$ 0.15 & 6.52 $\pm$ 0.35 & 8.30 $\pm$ 0.03 & 7.63 $\pm$ 0.34 \\
Multi-Head & 6.48 $\pm$ 0.33 & 6.55 $\pm$ 0.76 & 11.25 $\pm$ 0.75 & 7.88 $\pm$ 0.05 & 6.23 $\pm$ 0.40 & 8.40 $\pm$ 0.00 & 7.30 $\pm$ 0.13 \\
Multi-Head + GradNorm & 8.14 $\pm$ 0.27 & 9.42 $\pm$ 0.32 & 10.00 $\pm$ 0.50 & 10.50 $\pm$ 0.43 & 13.25 $\pm$ 0.50 & 10.10 $\pm$ 0.70 & 9.71 $\pm$ 0.42 \\
Multi-Head + PcGrad & 8.52 $\pm$ 0.30 & 7.76 $\pm$ 0.24 & 9.50 $\pm$ 2.00 & 8.39 $\pm$ 0.61 & 10.75 $\pm$ 0.12 & 7.93 $\pm$ 0.13 & 8.31 $\pm$ 0.32 \\
\bottomrule
\end{tabular}}\label{tab:avg-rank-32-100}
\end{table}

\begin{table}[htbp]
\centering
\caption{Hypervolume results for synthetic functions with 32 solutions and 100th percentile evaluations. For each task, algorithms within one standard deviation of having the highest performance are \textbf{bolded}.}
% \usepackage{booktabs}
\resizebox{\linewidth}{!}{
\begin{tabular}{c|ccccccccccccccccccc}
\toprule
Methods  & DTLZ1 & DTLZ2 & DTLZ3 & DTLZ4 & DTLZ5 & DTLZ6 & DTLZ7 & OmniTest & VLMOP1 & VLMOP2 & VLMOP3 & ZDT1 & ZDT2 & ZDT3 & ZDT4 & ZDT6 \\

\midrule
$\mathcal{D}$(best) & 10.43 & 9.43 & 9.71 & \textbf{10.76} & 9.06 & 8.20 & 8.32 & 3.87 & 0.08 & 1.64 & 45.14 & 4.04 & 4.70 & 5.05 & \textbf{5.46} & 4.76 \\
End-to-End & 10.64 $\pm$ 0.00 & \textbf{10.65 $\pm$ 0.00} & 10.64 $\pm$ 0.00 & \textbf{10.76 $\pm$ 0.00} & \textbf{10.64 $\pm$ 0.01} & \textbf{10.62 $\pm$ 0.03} & 10.60 $\pm$ 0.09 & 4.35 $\pm$ 0.00 & 2.54 $\pm$ 2.25 & 4.05 $\pm$ 0.05 & 46.90 $\pm$ 0.01 & 2.70 $\pm$ 0.00 & 3.19 $\pm$ 0.01 & 5.33 $\pm$ 0.03 & 3.72 $\pm$ 0.36 & \textbf{4.90 $\pm$ 0.00} \\
End-to-End + GradNorm & 10.63 $\pm$ 0.01 & 10.64 $\pm$ 0.01 & \textbf{10.65 $\pm$ 0.00} & \textbf{10.76 $\pm$ 0.00} & \textbf{10.54 $\pm$ 0.09} & 10.64 $\pm$ 0.00 & \textbf{10.71 $\pm$ 0.00} & 3.76 $\pm$ 0.03 & 1.26 $\pm$ 1.08 & 2.79 $\pm$ 1.34 & 42.23 $\pm$ 0.98 & 4.77 $\pm$ 0.01 & 5.63 $\pm$ 0.02 & 5.22 $\pm$ 0.02 & 3.23 $\pm$ 0.03 & 4.73 $\pm$ 0.11 \\
End-to-End + PcGrad & 10.63 $\pm$ 0.01 & 10.57 $\pm$ 0.04 & \textbf{10.65 $\pm$ 0.00} & \textbf{10.76 $\pm$ 0.00} & 9.02 $\pm$ 0.10 & 9.45 $\pm$ 0.15 & 10.52 $\pm$ 0.00 & 4.35 $\pm$ 0.00 & 4.80 $\pm$ 0.02 & 4.06 $\pm$ 0.01 & 46.79 $\pm$ 0.06 & \textbf{4.84 $\pm$ 0.01} & 5.70 $\pm$ 0.01 & 5.45 $\pm$ 0.00 & 3.12 $\pm$ 0.01 & 2.04 $\pm$ 0.22 \\
Multi-Head & 10.37 $\pm$ 0.26 & \textbf{10.65 $\pm$ 0.00} & 10.64 $\pm$ 0.00 & \textbf{10.76 $\pm$ 0.00} & \textbf{10.61 $\pm$ 0.03} & 10.65 $\pm$ 0.00 & 10.52 $\pm$ 0.00 & 4.24 $\pm$ 0.00 & \textbf{2.57 $\pm$ 2.25} & 4.08 $\pm$ 0.01 & 46.93 $\pm$ 0.00 & 2.72 $\pm$ 0.00 & \textbf{5.72 $\pm$ 0.00} & 5.41 $\pm$ 0.10 & 3.70 $\pm$ 0.23 & 4.87 $\pm$ 0.01 \\
\bottomrule
\end{tabular}}\label{tab:syn-32-100}
\end{table}

\begin{table}[htbp]
\centering
\caption{Hypervolume results for MO-NAS with 32 solutions and 100th percentile evaluations. For each task, algorithms within one standard deviation of having the highest performance are \textbf{bolded}.}
% \usepackage{booktabs}
\resizebox{\linewidth}{!}{
\begin{tabular}{c|ccccccccccccccccccc}
\toprule
Methods  & C-10/MOP1 & C-10/MOP2 & C-10/MOP3 & C-10/MOP4 & C-10/MOP5 & C-10/MOP6 & C-10/MOP7 & C-10/MOP8 & C-10/MOP9 & IN-1K/MOP1 & IN-1K/MOP2 & IN-1K/MOP3 & IN-1K/MOP4 & IN-1K/MOP5 & IN-1K/MOP6 & IN-1K/MOP7 & IN-1K/MOP8 & IN-1K/MOP9 & NasBench201-Test \\

\midrule
$\mathcal{D}$(best) & 4.78 & 10.48 & 9.72 & 21.15 & 40.51 & 92.43 & 358.27 & 4.55 & 10.59 & 4.97 & 5.00 & 11.21 & 16.63 & 17.26 & 44.43 & 4.69 & 11.42 & 21.50 & 10.07 \\
End-to-End & 4.82 $\pm$ 0.00 & 10.22 $\pm$ 0.26 & 10.63 $\pm$ 0.00 & 24.21 $\pm$ 0.12 & 49.48 $\pm$ 0.00 & 109.05 $\pm$ 0.00 & \textbf{512.31 $\pm$ 1.40} & 4.61 $\pm$ 0.07 & 10.68 $\pm$ 0.00 & 4.80 $\pm$ 0.02 & 5.11 $\pm$ 0.00 & 10.93 $\pm$ 0.06 & 16.88 $\pm$ 0.04 & 17.68 $\pm$ 0.03 & 45.39 $\pm$ 0.33 & 5.07 $\pm$ 0.07 & 11.07 $\pm$ 0.04 & 22.38 $\pm$ 0.37 & 10.28 $\pm$ 0.01 \\
End-to-End + GradNorm & 4.83 $\pm$ 0.00 & 10.49 $\pm$ 0.03 & 10.45 $\pm$ 0.06 & 24.96 $\pm$ 0.30 & \textbf{50.35 $\pm$ 0.02} & 108.57 $\pm$ 0.41 & 508.72 $\pm$ 0.60 & 4.55 $\pm$ 0.01 & 10.26 $\pm$ 0.17 & 5.14 $\pm$ 0.00 & 4.88 $\pm$ 0.03 & 10.62 $\pm$ 0.07 & \textbf{16.87 $\pm$ 0.16} & 17.49 $\pm$ 0.25 & 44.38 $\pm$ 0.63 & 4.97 $\pm$ 0.00 & 10.44 $\pm$ 0.06 & 22.71 $\pm$ 0.19 & 9.39 $\pm$ 0.32 \\
End-to-End + PcGrad & 4.83 $\pm$ 0.00 & 10.50 $\pm$ 0.00 & 10.53 $\pm$ 0.06 & 25.30 $\pm$ 0.15 & 50.17 $\pm$ 0.01 & 108.22 $\pm$ 0.52 & 501.11 $\pm$ 0.00 & 4.34 $\pm$ 0.18 & \textbf{10.93 $\pm$ 0.27} & 5.01 $\pm$ 0.17 & 5.05 $\pm$ 0.01 & 11.14 $\pm$ 0.13 & 16.16 $\pm$ 0.61 & 17.44 $\pm$ 0.04 & 44.07 $\pm$ 0.26 & 4.93 $\pm$ 0.00 & 10.95 $\pm$ 0.01 & 23.39 $\pm$ 0.00 & \textbf{10.32 $\pm$ 0.02} \\
\bottomrule
\end{tabular}}\label{tab:nas-32-100}
\end{table}

\begin{table}[htbp]
\centering
\caption{Hypervolume results for MORL with 32 solutions and 100th percentile evaluations. For each task, algorithms within one standard deviation of having the highest performance are \textbf{bolded}.}
\begin{tabular}{c|ccccccccccccccccccc}
\toprule
Methods & MO-Hopper & MO-Swimmer \\

\midrule
$\mathcal{D}$(best) & 4.21 & \textbf{2.85} \\
End-to-End & 4.04 $\pm$ 0.24 & 2.56 $\pm$ 0.13 \\
End-to-End + GradNorm & 4.42 $\pm$ 0.36 & 2.61 $\pm$ 0.03 \\
End-to-End + PcGrad & 4.51 $\pm$ 0.27 & 2.71 $\pm$ 0.05 \\
Multi-Head & 3.81 $\pm$ 0.04 & 2.36 $\pm$ 0.13 \\
Multi-Head + GradNorm & 3.71 $\pm$ 0.03 & 2.59 $\pm$ 0.04 \\
Multi-Head + PcGrad & 4.40 $\pm$ 0.24 & 2.18 $\pm$ 0.19 \\
Multiple Models & 3.87 $\pm$ 0.00 & 2.28 $\pm$ 0.11 \\
Multiple Models + COMs & \textbf{4.90 $\pm$ 0.06} & \textbf{2.85 $\pm$ 0.04} \\
Multiple Models + RoMA & 4.18 $\pm$ 0.03 & \textbf{2.67 $\pm$ 0.18} \\
Multiple Models + IOM & \textbf{4.63 $\pm$ 0.33} & 2.23 $\pm$ 0.02 \\
Multiple Models + ICT & 3.95 $\pm$ 0.03 & 2.69 $\pm$ 0.02 \\
Multiple Models + Tri-Mentoring & 3.70 $\pm$ 0.01 & 2.41 $\pm$ 0.02 \\
MOBO & 4.20 $\pm$ 0.01 & 2.27 $\pm$ 0.10 \\
MOBO-$q$ParEGO & N/A & N/A \\
MOBO-JES & N/A & N/A \\ \bottomrule
\end{tabular}\label{tab:morl-32-100}
\end{table}

\begin{table}[htbp]
\centering
\caption{Hypervolume results for MOCO with 32 solutions and 100th percentile evaluations. For each task, algorithms within one standard deviation of having the highest performance are \textbf{bolded}.}
% \usepackage{booktabs}
\resizebox{\linewidth}{!}{
\begin{tabular}{c|ccccccccccccccccccc}
\toprule
Methods & Bi-CVRP-20 & Bi-CVRP-50 & Bi-CVRP-100 & Bi-KP-50 & Bi-KP-100 & Bi-KP-200 & Bi-TSP-20 & Bi-TSP-50 & Bi-TSP-100 & Bi-TSP-500 & Tri-TSP-20 & Tri-TSP-50 & Tri-TSP-100 & Portfolio\\

\midrule
$\mathcal{D}$(best) & \textbf{5.37} & \textbf{5.11} & \textbf{4.93} & \textbf{3.00} & \textbf{3.45} & \textbf{4.68} & \textbf{5.05} & \textbf{4.89} & \textbf{4.55} & \textbf{4.52} & \textbf{11.88} & \textbf{9.82} & \textbf{9.36} & \textbf{3.78} \\
End-to-End & 5.05 $\pm$ 0.17 & 4.30 $\pm$ 0.12 & 4.53 $\pm$ 0.36 & 2.70 $\pm$ 0.04 & 2.33 $\pm$ 0.30 & 3.37 $\pm$ 0.00 & 3.44 $\pm$ 0.03 & 4.68 $\pm$ 0.12 & 4.48 $\pm$ 0.02 & \textbf{4.48 $\pm$ 0.05} & 6.87 $\pm$ 1.32 & 5.56 $\pm$ 0.70 & 9.07 $\pm$ 0.04 & 2.95 $\pm$ 0.01 \\
End-to-End + GradNorm & 4.02 $\pm$ 0.27 & 3.75 $\pm$ 0.21 & 3.63 $\pm$ 0.00 & 2.67 $\pm$ 0.01 & 2.67 $\pm$ 0.02 & 2.69 $\pm$ 0.00 & 3.06 $\pm$ 0.30 & 3.87 $\pm$ 0.13 & 3.53 $\pm$ 0.00 & 4.11 $\pm$ 0.16 & 5.61 $\pm$ 0.01 & 6.29 $\pm$ 0.00 & 8.47 $\pm$ 0.12 & 3.13 $\pm$ 0.00 \\
End-to-End + PcGrad & 3.59 $\pm$ 0.17 & 3.52 $\pm$ 0.40 & 3.63 $\pm$ 0.12 & 2.75 $\pm$ 0.00 & 2.58 $\pm$ 0.22 & 2.84 $\pm$ 0.15 & 3.14 $\pm$ 0.24 & 4.49 $\pm$ 0.02 & 4.18 $\pm$ 0.17 & 3.98 $\pm$ 0.02 & 9.31 $\pm$ 0.08 & 8.11 $\pm$ 0.00 & 8.54 $\pm$ 0.18 & 3.02 $\pm$ 0.00 \\
Multi-Head & 4.65 $\pm$ 0.00 & 3.75 $\pm$ 0.12 & 4.88 $\pm$ 0.01 & 2.73 $\pm$ 0.03 & 2.45 $\pm$ 0.14 & 2.99 $\pm$ 0.24 & 3.10 $\pm$ 0.14 & 4.50 $\pm$ 0.02 & 3.48 $\pm$ 0.32 & 3.19 $\pm$ 0.04 & 5.14 $\pm$ 0.05 & 6.65 $\pm$ 1.16 & 8.34 $\pm$ 0.32 & 2.98 $\pm$ 0.16 \\
Multi-Head + GradNorm & 3.77 $\pm$ 0.33 & 3.21 $\pm$ 0.07 & 3.04 $\pm$ 0.17 & 2.29 $\pm$ 0.16 & 2.24 $\pm$ 0.22 & 1.64 $\pm$ 0.00 & 3.69 $\pm$ 0.08 & 3.23 $\pm$ 0.01 & 3.27 $\pm$ 0.17 & 3.18 $\pm$ 0.08 & 5.96 $\pm$ 0.47 & 5.53 $\pm$ 0.30 & 6.73 $\pm$ 0.00 & 3.11 $\pm$ 0.11 \\
\bottomrule
\end{tabular}}\label{tab:moco-32-100}
\end{table}

\begin{table}[htbp]
\centering
\caption{Hypervolume results for scientific design with 32 solutions and 100th percentile evaluations. For each task, algorithms within one standard deviation of having the highest performance are \textbf{bolded}.}
% \usepackage{booktabs}

\begin{tabular}{c|ccccccccccccccccccc}
\toprule
Methods  & Molecule & Regex & RFP & ZINC \\

\midrule
$\mathcal{D}$(best) & 2.26 & 3.05 & 3.75 & 4.06 \\
End-to-End & 1.95 $\pm$ 0.13 & 2.80 $\pm$ 0.00 & \textbf{4.34 $\pm$ 0.03} & \textbf{4.12 $\pm$ 0.01} \\
End-to-End + GradNorm & 1.05 $\pm$ 0.00 & 2.38 $\pm$ 0.00 & 3.46 $\pm$ 0.08 & 4.00 $\pm$ 0.01 \\
End-to-End + PcGrad & \textbf{3.33 $\pm$ 0.00} & 2.47 $\pm$ 0.00 & 3.62 $\pm$ 0.00 & 4.04 $\pm$ 0.04 \\
Multi-Head & 2.20 $\pm$ 0.38 & 2.80 $\pm$ 0.00 & 3.67 $\pm$ 0.00 & \textbf{4.12 $\pm$ 0.00} \\
Multi-Head + GradNorm & 1.06 $\pm$ 0.05 & 2.38 $\pm$ 0.00 & 3.67 $\pm$ 0.00 & 3.82 $\pm$ 0.00 \\
Multi-Head + PcGrad & 1.04 $\pm$ 0.00 & 2.47 $\pm$ 0.00 & 3.71 $\pm$ 0.00 & 3.92 $\pm$ 0.00 \\
Multiple Models & 1.19 $\pm$ 0.14 & 2.38 $\pm$ 0.00 & 3.70 $\pm$ 0.00 & 4.01 $\pm$ 0.01 \\
Multiple Models + COMs & 2.16 $\pm$ 0.09 & 2.05 $\pm$ 0.00 & 3.65 $\pm$ 0.00 & 3.97 $\pm$ 0.10 \\
Multiple Models + RoMA & 1.84 $\pm$ 0.10 & 2.47 $\pm$ 0.00 & 3.71 $\pm$ 0.00 & 3.95 $\pm$ 0.00 \\
Multiple Models + IOM & 1.72 $\pm$ 0.36 & 2.47 $\pm$ 0.00 & \textbf{4.06 $\pm$ 0.35} & 4.08 $\pm$ 0.03 \\
Multiple Models + ICT & \textbf{2.65 $\pm$ 1.31} & 2.47 $\pm$ 0.00 & 3.76 $\pm$ 0.00 & 4.05 $\pm$ 0.02 \\
Multiple Models + Tri-Mentoring & 2.03 $\pm$ 0.00 & 2.47 $\pm$ 0.00 & 3.75 $\pm$ 0.00 & 3.94 $\pm$ 0.00 \\
MOBO & 1.15 $\pm$ 0.02 & \textbf{4.52 $\pm$ 0.18} & 3.69 $\pm$ 0.01 & \textbf{4.18 $\pm$ 0.08} \\
MOBO-$q$ParEGO & 2.12 $\pm$ 0.04 & 4.26 $\pm$ 0.25 & 3.33 $\pm$ 0.00 & 4.05 $\pm$ 0.02 \\
MOBO-JES & 2.10 $\pm$ 1.04 & N/A & N/A & N/A \\\bottomrule
\end{tabular}\label{tab:sci-design-32-100}
\end{table}

\begin{table}[htbp]
\centering
\caption{Hypervolume results for RE with 32 solutions and 100th percentile evaluations. For each task, algorithms within one standard deviation of having the highest performance are \textbf{bolded}.}
% \usepackage{booktabs}
\resizebox{\linewidth}{!}{
\begin{tabular}{c|ccccccccccccccccccc}
\toprule

Methods & RE21 & RE22 & RE23 & RE24 & RE25 & RE31 & RE32 & RE33 & RE34 & RE35 & RE36 & RE37 & RE41 & RE42 & RE61 \\

\midrule
$\mathcal{D}$(best) & 4.23 & 4.78 & 4.75 & 4.59 & 4.79 & 10.23 & 10.53 & 10.59 & 48.06 & 10.96 & 7.57 & 4.72 & 36.17 & 12.53 & 135.87 \\
End-to-End & 4.42 $\pm$ 0.00 & 4.84 $\pm$ 0.00 & \textbf{4.84 $\pm$ 0.00} & 4.38 $\pm$ 0.00 & 4.82 $\pm$ 0.02 & 10.58 $\pm$ 0.03 & 10.56 $\pm$ 0.08 & 10.68 $\pm$ 0.00 & 52.82 $\pm$ 0.05 & 11.61 $\pm$ 0.01 & 8.24 $\pm$ 0.00 & 6.20 $\pm$ 0.00 & 42.53 $\pm$ 0.05 & 17.09 $\pm$ 2.14 & 140.13 $\pm$ 1.42 \\
End-to-End + GradNorm & 4.81 $\pm$ 0.01 & 4.84 $\pm$ 0.00 & 2.64 $\pm$ 0.00 & 4.38 $\pm$ 0.00 & \textbf{4.84 $\pm$ 0.00} & 10.65 $\pm$ 0.00 & 10.63 $\pm$ 0.00 & 9.90 $\pm$ 0.00 & 51.98 $\pm$ 0.63 & 11.51 $\pm$ 0.00 & 8.63 $\pm$ 0.71 & 6.11 $\pm$ 0.10 & 39.21 $\pm$ 0.12 & 13.46 $\pm$ 0.00 & 140.24 $\pm$ 0.25 \\
End-to-End + PcGrad & 4.86 $\pm$ 0.01 & 4.84 $\pm$ 0.00 & 4.84 $\pm$ 0.00 & 4.40 $\pm$ 0.00 & 4.35 $\pm$ 0.00 & \textbf{10.65 $\pm$ 0.00} & 10.65 $\pm$ 0.00 & 10.41 $\pm$ 0.07 & 52.78 $\pm$ 0.04 & 11.66 $\pm$ 0.01 & 9.76 $\pm$ 0.11 & 5.52 $\pm$ 0.00 & 41.64 $\pm$ 0.31 & 15.92 $\pm$ 1.75 & 139.60 $\pm$ 0.28 \\
Multi-Head & 4.61 $\pm$ 0.13 & 4.84 $\pm$ 0.00 & 4.70 $\pm$ 0.04 & 4.78 $\pm$ 0.00 & 4.35 $\pm$ 0.00 & 10.65 $\pm$ 0.00 & 10.64 $\pm$ 0.00 & 10.65 $\pm$ 0.01 & 52.63 $\pm$ 0.22 & 11.70 $\pm$ 0.00 & 6.76 $\pm$ 0.00 & 5.72 $\pm$ 0.00 & 42.45 $\pm$ 0.40 & 19.30 $\pm$ 0.32 & 141.50 $\pm$ 0.03 \\
Multi-Head + GradNorm & 4.91 $\pm$ 0.00 & 4.65 $\pm$ 0.12 & 4.83 $\pm$ 0.00 & 3.51 $\pm$ 0.87 & 4.31 $\pm$ 0.53 & 10.65 $\pm$ 0.00 & 10.63 $\pm$ 0.00 & 5.82 $\pm$ 0.02 & 52.81 $\pm$ 0.00 & 11.51 $\pm$ 0.00 & 0.02 $\pm$ 0.00 & 6.26 $\pm$ 0.01 & 42.56 $\pm$ 0.11 & 15.82 $\pm$ 2.36 & 140.23 $\pm$ 0.66 \\
\bottomrule
\end{tabular}}\label{tab:re-32-100}
\end{table}

% \clearpage
% \newpage

% \begin{table}[htbp]
% \centering
% \caption{Running Time}\label{table:running time}
% % Please add the following required packages to your document preamble:
% % \usepackage{booktabs}
% \begin{tabular}{c|ccccccccccccccccccc}
% \toprule
% Task type       & Synthetic & MO-NAS    & MORL      & MOCO      & Sci-Design & RE     \\
% \midrule
% Task                   & VLMOP2    & C-10/MOP3 & MO-Hopper & Bi-TSP-20 & Regex      & RE21   \\
% \midrule

% End-to-End + GradNorm  & 444/21    & 216/4     & 93/200    & 536/22    & 316/10     & 465/20 \\
% End-to-End + PcGrad    & 416/20    & 209/4     & 120/204   & 467/22    & 326/9      & 469/21 \\

% Multi-Head + GradNorm  & 524/20    & 324/4     & 98/210    & 627/22    & 440/10     & 664/21 \\
% Multi-Head + PcGrad    & 506/20    & 244/4     & 121/180   & 550/22    & 389/14     & 409/21 \\

% Multiple Models + COMs & 706/21    & 318/3     & 114/179   & 919/23    & 492/8      & 690/22 \\
% MOBO                   & 2/247     & 1/4       & 7/1214    & 1/144     & 1/1        & 1/374 \\ \bottomrule
% \end{tabular}
% \end{table}

% \usepackage{booktabs}
% \usepackage{booktabs}
% \usepackage{booktabs}

\subsection{Additional Visualization Results}
\begin{figure}[t!]\centering
\includegraphics[width=1\linewidth]{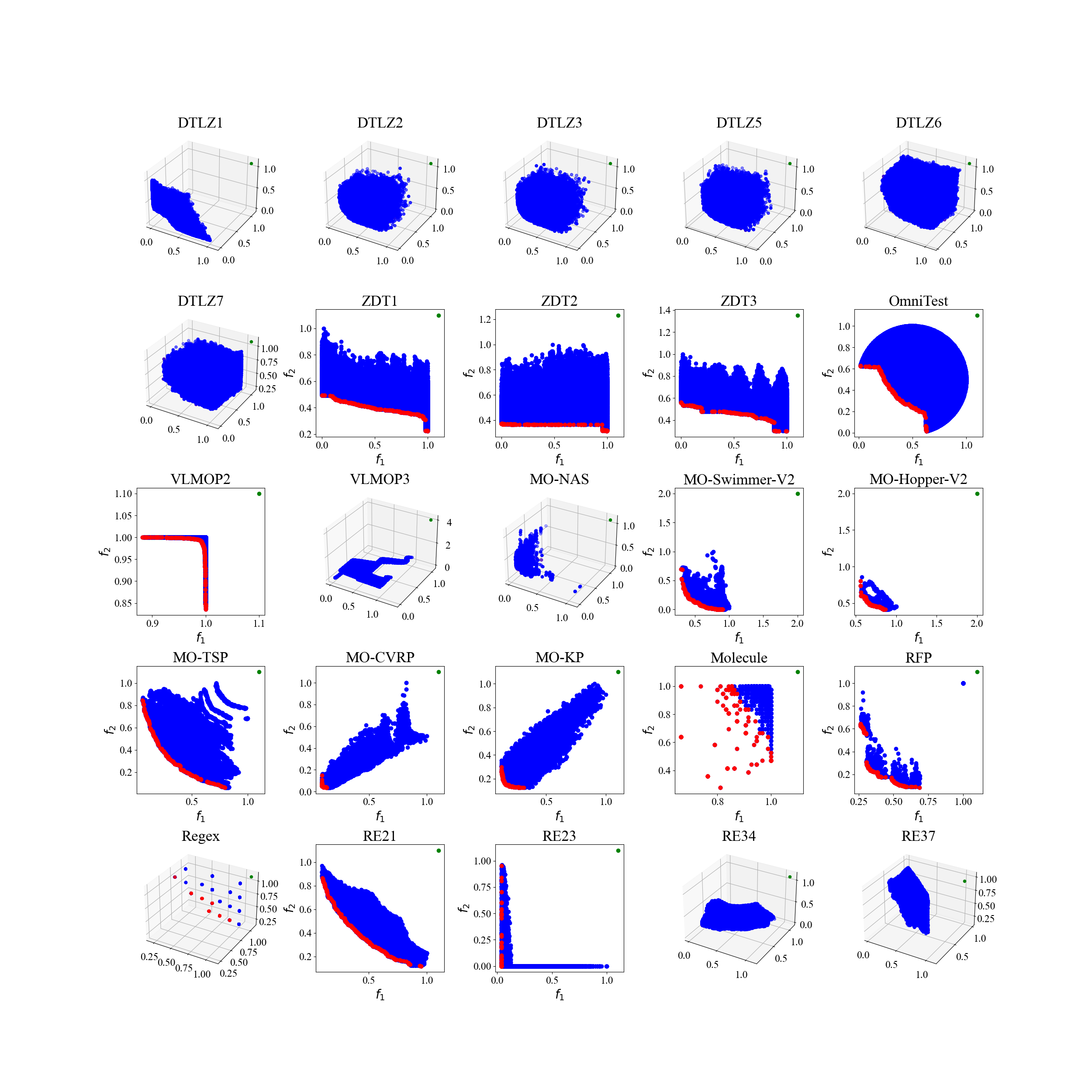}
\caption{Visualization of datasets in Off-MOO-Bench. Blue points represent the offline dataset, and red points represent the 256 best-non-dominated solutions over the dataset. Note that some red dots are not visible in the graph due to the plot perspective.}\label{fig:dataset-all}
\end{figure}

\begin{figure}[t!]\centering
\includegraphics[width=1\linewidth]{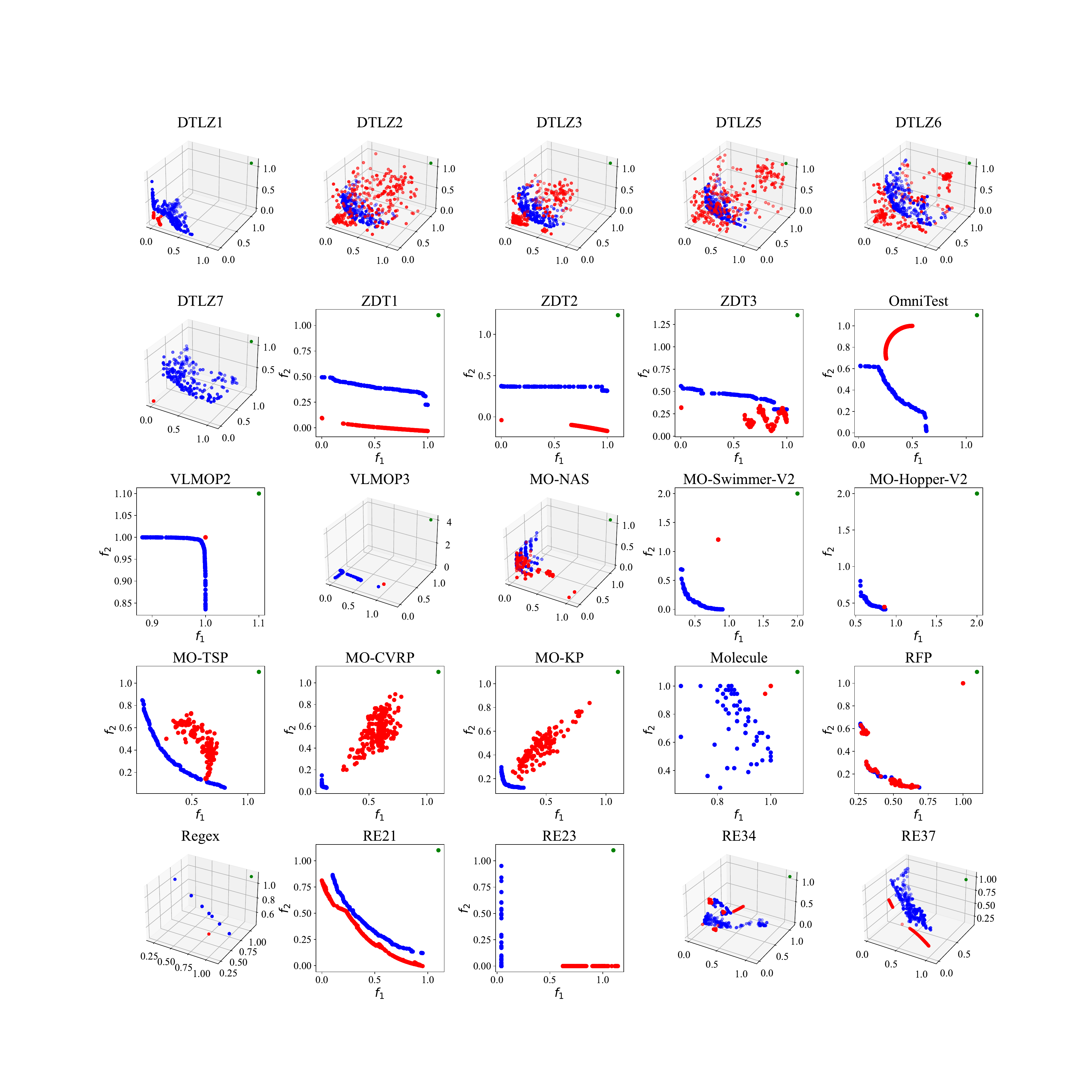}
\caption{Visualization of Pareto fronts found by Multi-Head + GradNorm. Blue points represent the initial population, which are the 256 best-non-dominated solutions over the offline dataset. Red points represent the Pareto fronts found by algorithm, and green points represent the reference points (i.e., nadir points) that are set by us manually.}\label{fig:pf-MH-GN}
\end{figure}

In this section, we visualize all the tasks with number of objectives less than 3, for better understanding the tasks. We first show the dataset of tasks in Off-MOO-Bench in Figure~\ref{fig:dataset-all}, and then show the Pareto fronts found by Multi-Head + GradNorm in Figure~\ref{fig:pf-MH-GN}.

% \begin{figure*}[tb]\centering
% \includegraphics[width=0.24\linewidth]{Fig/pf_rfp.png}
% \includegraphics[width=0.24\linewidth]{Fig/pf_rfp.png}
% \includegraphics[width=0.24\linewidth]{Fig/pf_rfp.png}
% \includegraphics[width=0.24\linewidth]{Fig/pf_rfp.png}
% \\
% \includegraphics[width=0.24\linewidth]{Fig/pf_rfp.png}
% \includegraphics[width=0.24\linewidth]{Fig/pf_rfp.png}
% \includegraphics[width=0.24\linewidth]{Fig/pf_rfp.png}
% \includegraphics[width=0.24\linewidth]{Fig/pf_rfp.png}
% \caption{Objective space visualization of some tasks in the dataset. Each colored point denotes the objective vector of a data point, where the green one is the reference point and the red one belongs to the Pareto front.}\label{fig:main-pareto-set}
% \end{figure*}

\end{document}